\def\year{2022}\relax
\documentclass[letterpaper]{article} 
\usepackage{aaai22}  
\usepackage{times}  
\usepackage{helvet}  
\usepackage{courier}  
\usepackage[hyphens]{url}  
\usepackage{graphicx} 
\usepackage{natbib}  
\usepackage{caption} 
\DeclareCaptionStyle{ruled}{labelfont=normalfont,labelsep=colon,strut=off} 
\usepackage{amssymb,amsmath,amsfonts}
\usepackage{setspace,multicol,multirow,enumerate,textcomp}
\usepackage{stackengine,calc,xspace,array}
\usepackage{verbatim}
\usepackage{hyperref}
\DeclareSymbolFont{rsfscript}{OMS}{rsfs}{m}{n}
\DeclareSymbolFontAlphabet{\mathrsfs}{rsfscript}
\usepackage[T1]{fontenc}
\DeclareSymbolFont{rsfscript}{OMS}{rsfs}{m}{n}
\DeclareSymbolFontAlphabet{\mathrsfs}{rsfscript}
\newtheorem{theorem}{Theorem}
\newtheorem{example}[theorem]{Example}

\makeatletter
\def\namedlabel#1#2{\begingroup
   \def\@currentlabel{#2}%
   \label{#1}\endgroup
}
\makeatother

\usepackage{algorithm,algorithmicx}
\usepackage[noend]{algpseudocode}
\MakeRobust{\Call}

\DeclareMathOperator*{\argmax}{arg\,max}

\usepackage[dvipsnames]{xcolor}

\usepackage[makeroom]{cancel}

\usepackage{algorithm,algorithmicx}
\usepackage[noend]{algpseudocode}
\MakeRobust{\Call}
\newcommand{\LineIf}[2]{\State \algorithmicif\ {#1}\ \algorithmicthen\ {#2}}

\newcommand{\ARXIVorAAAI}[2]{#1}

\newcommand{\orthodox}{$\mathbb{O}$}

\newcommand{\orthodoxMR}{$\mathbb{O}^\text{tree:RAVE}_\text{sim:MAST}$\xspace}
\newcommand{\semisplit}[3]{
  $\mathbb{S}^\text{tree:#1}_\text{sim:#2}\text{\footnotesize@#3}$}
\newcommand{\semisplitV}[3]{
  \setlength\tabcolsep{0cm}
  \normalsize\begin{tabular}{l}
  $\mathbb{S}^\text{tree:#1}_\text{sim:#2}$\\
  \phantom{-}\\
  \scriptsize\phantom{$\mathbb{S}$}@#3\end{tabular}
}

\usepackage{lipsum}
\usepackage{listings}
\lstdefinelanguage{RBG}{
	alsoletter  = {\#},
    basicstyle=\small,
    stepnumber=1,
    numbers=left,
    numbersep=8pt,
    xleftmargin=16pt,
    numberstyle=\color{gray},
    keywords={\#board,\#variables,\#pieces,\#players,\#rules,
      \#name,\#bPieces,\#diagonalMove,\#wBishopMove,\#wQueenMove,
      \#anySquare,\#turn,\#line,\#m0,\#m1,\#m2,\#m3,\#m4,\#m5,\#m6,\#m7,\#pickUpPiece,\#basicMove,\#endGame,\#checkForWin,\#fullMove,\#up8Times,\#majorPieces,\#promotePawn,\#directedShift,\#queenShift
    },
    keywordstyle=\bfseries\color{purple},
    keywords=[2]{
        and,or,not},
    keywordstyle=[2]\bfseries\color{black},
    morecomment=[l]{//}, 
    morecomment=[s]{/*}{*/}, 
    commentstyle=\itshape\color{cyan},
    morestring=[b]", 
    stringstyle=\color{teal},
}
\newcommand{\orthodoxM}{$\mathbb{O}_\text{sim:MAST}$\xspace}
\newcommand{\orthodoxR}{$\mathbb{O}^\text{tree:RAVE}$\xspace}

\usepackage{tikz}
\usetikzlibrary{arrows}
\usetikzlibrary{calc}
\usetikzlibrary{fit}
\usetikzlibrary{quotes}
\usetikzlibrary{shapes}

\newcommand{\mastsplit}{MAST-split\xspace}
\newcommand{\mastcontext}{MAST-context\xspace}
\newcommand{\mastjoin}{MAST-join\xspace}
\newcommand{\mastmix}{MAST-mix\xspace}
\newcommand{\ravesplit}{RAVE-split\xspace}
\newcommand{\ravecontext}{RAVE-context\xspace}
\newcommand{\ravejoin}{RAVE-join\xspace}
\newcommand{\ravemix}{RAVE-mix\xspace}

\title{Split Moves for Monte-Carlo Tree Search}
\author{Jakub Kowalski\textsuperscript{\rm 1}, Maksymilian Mika\textsuperscript{\rm 1}, Wojciech Pawlik\textsuperscript{\rm 1}, Jakub Sutowicz\textsuperscript{\rm 1}, Marek Szyku{\l}a\textsuperscript{\rm 1}, \\Mark H. M. Winands\textsuperscript{\rm 2}}

\affiliations {
\textsuperscript{\rm 1}University of Wroc{\l}aw, Faculty of Mathematics and Computer Science\\
\textsuperscript{\rm 2}Maastricht University, Department of Data Science and Knowledge Engineering\\
jko@cs.uni.wroc.pl, mika.maksymilian@gmail.com, pawlik.wj@gmail.com, \\jakubsutowicz@gmail.com, msz@cs.uni.wroc.pl, m.winands@maastrichtuniversity.nl
}

\begin{document} 

\maketitle 
\thispagestyle{plain}
\pagestyle{plain}

\begin{abstract}
In many games, moves consist of several decisions made by the player. These decisions can be viewed as separate moves, which is already a common practice in multi-action games for efficiency reasons. Such division of a player move into a sequence of simpler / lower level moves is called \emph{splitting}. So far, split moves have been applied only in forementioned straightforward cases, and furthermore, there was almost no study revealing its impact on agents' playing strength. Taking the knowledge-free perspective, we aim to answer how to effectively use split moves within Monte-Carlo Tree Search (MCTS) and what is the practical impact of split design on agents' strength. This paper proposes a generalization of MCTS that works with arbitrarily split moves. We design several variations of the algorithm and try to measure the impact of split moves separately on efficiency, quality of MCTS, simulations, and action-based heuristics. The tests are carried out on a set of board games and performed using the Regular Boardgames General Game Playing formalism, where split strategies of different granularity can be automatically derived based on an abstract description of the game. The results give an overview of the behavior of agents using split design in different ways. We conclude that split design can be greatly beneficial for single- as well as multi-action games.
\end{abstract}

\section{Introduction}

The benefits of simulation-based, knowledge-free, open-loop algorithms such as Monte-Carlo Tree Search (MCTS) \cite{Kocsis2006Bandit,Browne2012ASurvey,swiechowski2021monte} and Rolling Horizon Evolutionary Algorithm \cite{perez2013rolling} are especially suited to work within environments with many unknowns.
In particular, they are widely used in General Game Playing (GGP) \cite{Genesereth2005General}, a domain focusing on developing agents that can successfully play any game given its formalized rules, which was established to promote work in generalized, practically applicable algorithms \cite{Finnsson2010Learning,thielscherAAAI20}.
Initially based entirely on Stanford's Game Description Language (GDL) \cite{Love2006General}, GGP expands over time as new game description formalisms are being developed e.g., Toss \cite{Kaiser2011FirstOrder}, GVG-AI \cite{Perez2016General}, Regular Boardgames (RBG) \cite{Kowalski2019RegularBoardgames}, and Ludii \cite{piette2019ludii}.

Recent advances in search and learning support the trend of generalization, focusing on methods being as widely applicable as possible.
Deep Q-networks were applied to play classic Atari games and achieved above human-level performance on most of the 49 games from the test set \cite{Mnih2015HumanLevel}. More recently, \textsc{AlphaZero}, showed how to utilize a single technique to play Go, Chess, and Shogi on a level above all other compared AI agents \cite{silver2018general}. 

In the trend of developing enhancements for MCTS \cite{cazenave2015grave,baier2018mcts}, we tackle the problem of influencing the quality of the search by altering the structure of the game tree itself.
In many games, a player's turn consists of a sequence of choices that can be examined separately.
A straightforward representation is to encode these choices as distinct moves, obtaining a split game tree, instead of using a single move in orthodox design.
The potential applications go beyond games, as the method can be used for any kind of problem that is solvable via a simulation-based approach and its representation of actions can be decomposed.  
In this paper, we are interested in the general technique of altering the game tree by introducing split moves and its possible effects, rather than its application to a particular game combined with expert knowledge.
Hence, we focus on the GGP setting and MCTS, which is the most well-known and widely applied general search algorithm working without expert knowledge.

We propose the \emph{semisplit} algorithm, which is a generalization of MCTS that effectively works with arbitrary splitting.
For the purposes of experiments, we implement the concept in the Regular Boardgames system \cite{Kowalski2019RegularBoardgames} -- a universal GGP formalism for the class of finite deterministic games with perfect information.
A few \emph{split strategies} of different granularity are developed, which split moves basing on the given general game description.
The experiments are conducted on a set of classic board games, comparing agents using orthodox and split designs.
We test a number of variants of the algorithm, applying split moves selectively to different phases of MCTS, and include action-based heuristics (MAST, RAVE \cite{gelly2007combining}) to observe the behavior also for enhanced MCTS.
From the results, we identify the most beneficial configurations and conclude that split moves can greatly improve the player's strength.

\ARXIVorAAAI{The source code used for the experiments is shared within the RBG implementation \cite{Kowalski2021RBGsource}.}{The full version of this paper is available at~\cite{Kowalski2021SplitMovesextended}, and the source code used for the experiments is shared within the RBG implementation \cite{Kowalski2021RBGsource}.}

\subsection{Related Work}

The idea of splitting moves is well known, but apparently, it was not given proper consideration in the literature, being either used naturally in trivial cases or restrained to follow a human-authored heuristic.
Even these cases were discussed in rather limited aspects, given how general applications of split technique can be.

For particularly complex environments, split is regarded as natural and mandatory.
This technique is widely used for Arimaa, Hearthstone, and other multi-action games \cite{Fotland06Building,justesen2017playing,roelofs2017pitfalls}.
Here, the reduced branching factor is considered to be the main effect, as otherwise, programs could not play such games at a proper level.
The case of Amazons is the only one that we have found where agents playing with split and non-split representations were compared against each other \cite{kloetzer2007monte}. 
For multi-action games such as real-time strategies, where the ordering of actions is unrestricted, Combinatorial Multi-armed Bandits algorithms are often employed \cite{Ontanon17CombinatorialMAB}.
Using actions separately as moves in the MCTS tree was also considered under the name of \emph{hierarchical expansion} \cite{roelofs2017pitfalls} and in context of factoring action space for MDPs \cite{geisser2020trial}.
So far, splitting was applied only for such multi-action games, where it is possible and natural to divide a turn into separate moves and process them like regular ones. 
Splitting / move decomposition should not be confused with the game decomposition \cite{hufschmitt2019exploiting}.

A practical application of splitting is found in GGP, where many games are manually (re)encoded in their split variants just to improve efficiency.
For example, some split versions of games like Amazons, Arimaa, variants of Draughts, or Pentago exist in GDL.
The same approach is taken in other systems like Ludii.
However, it is generally unknown how such versions affect the agents' playing strength.

Additionally, splitting via artificial turns causes some repercussions, e.g., for game statistics or handling the turn timer.
Especially in a competitive setting, an agent gets the same time for every move, even when they are single actions, thus making more moves in a turn gives longer computation time.
Although this can be resolved, it requires specific language constructions that do not exist in any GGP system.
Instead of complicating languages, it would be better and more general to develop split-handling on the agent's side.

A related topic is \emph{move groups} \cite{saito2007grouping,childs2008transpositions,Eyck2011revisiting},
where during MCTS expansion, children of every tree node are partitioned into a constant number of classes guided by a heuristic.
The basic idea of move groups is to divide nodes of the MCTS tree into two levels, creating intermediate nodes that group children belonging to the same class.
There is no reported application of move groups beyond Go, Settlers of Catan, and artificial single-player game trees for maximizing UCT payoff. 
Usually, move groups are understood as introducing artificial tree nodes unrelated to any move representation nor simplifying computation.
They also require human intervention to encode rules on how to partition the moves.
From the perspective of splitting, (nested) move groups are a side effect, but not every partition can be obtained by splitting.

As splitting just alters the game tree, it is compatible with any other search algorithm that uses this game tree as the underlying structure. 
In particular, it affects action-based heuristics operating on moves such as MAST and RAVE \cite{gelly2007combining}. 
But unlike usual techniques, splitting often improves efficiency, so it would be beneficial assuming that it leaves the behavior of algorithms unchanged.

\section{Semisplitting in MCTS}

\subsection{Abstract Game}

We adapt a standard definition of an abstract turn-based game \cite{Rasmusen1994Games} to our goals.
A \emph{finite deterministic turn-based game with perfect information} (simply called \emph{game}) $\mathcal{G}$ is a tuple $(\mathit{players}_\mathcal{G},\mathcal{T}_\mathcal{G},\mathit{control}_\mathcal{G},\mathit{out}_\mathcal{G})$, where:
$\mathit{players}_\mathcal{G}$ is a finite non-empty set of \emph{players};
$\mathcal{T}_\mathcal{G}=(V,E)$ is a finite directed tree called the \emph{game tree}, where $V$ is the set of nodes called \emph{game states} and $E$ is the set of edges called \emph{moves}, $V_\mathrm{n} \subset V$ is the set of inner nodes called \emph{non-terminal} states, and $V_\mathrm{t} \subseteq V$ is the set of leaves called \emph{terminal} states;
$\mathit{control}_\mathcal{G}\colon V_\mathrm{n} \to \mathit{players}_\mathcal{G}$ is a function assigning the \emph{current player} to non-terminal states;
$\mathit{out}_\mathcal{G}\colon V_\mathrm{t} \times \mathit{players}_\mathcal{G} \to \mathbb{R}$ is a function assigning the final \emph{score} of each player at terminal states.
For a non-terminal state $s \in V_\mathrm{n}$, the set of \emph{legal moves} is the set of outgoing edges $\{(s,t) \in E \mid t \in V\}$. 
During a play, the current player $\mathit{control}_\mathcal{G}(s)$ chooses one of its legal moves.
The game tree is directed outwards its root $s_0 \in V$, which is called the \emph{initial state}.
Hence, all states are reachable from $s_0$.
Each play starts from $s_0$ and ends at a terminal state (leaf).

Two games are \emph{isomorphic} if there exists a bijection between the states that preserve the edges of the game tree and the current player (if the state is non-terminal) or the scores (if the state is terminal).
For a state $s \in V$, the \emph{subgame} $\mathcal{G}_s$ is the game with the tree obtained from $\mathcal{T}_\mathcal{G}$ by rooting at $s$ and removing all states unreachable from $s$.

\subsection{Semisplit Game}

Going deeper into a particular representation of a move, it usually can be partitioned into a sequence of smaller pieces, which we call \emph{semimoves}.
For example, depending on a particular implementation, they could correspond to atomic actions, move groups, or, in the extreme case, even single bits of a technical move representation.
Computing semimoves can be, but not always is, computationally easier than full moves and sometimes may reveal structural information desirable in a knowledge-based analysis.

However, often a natural and the most effective splitting does not lead to a proper variant of the game, i.e., we cannot treat semimoves as usual moves.
It is because not every available sequence of easily computed semimoves can be completed up to a legal move.
This especially concerns single-action games, but also splits inferred automatically in general, where without prior knowledge it is difficult to determine if we obtain a proper game.

\begin{example}
In Chess, a typical move consists of picking up a piece and choosing its destination square.
It would be much more efficient first to select a piece from the list of pieces and then a square from the list of available destinations computed just for this piece, than to select a move from the list of all legal ones, which is usually much longer.
However, sometimes we may not be able to make a legal move with a selected piece, e.g., because the piece is blocked or the king will be left under check.
\end{example}

A remedy could be checking if each available semimove is a prefix of at least one legal move.
However, in many cases, this can be as costly as computing all legal moves, losing performance benefits, or even decreasing efficiency.
Instead, we can work on semisplit games directly.

To provide an abstract model, we require a different game definition, including additional information about intermediate states.
We extend the definition of a game to a \emph{semisplit game} as follows.
Let $V$ be now the disjoint union of non-terminal states $V_\mathrm{n}$, terminal states $V_\mathrm{t}$, and the \emph{intermediate states} $V_\mathrm{i}$.
Then, $V_\mathrm{n}$ is a subset of inner vertices of the game tree, terminal states $V_\mathrm{t}$ is a subset of leaves, and $V_\mathrm{i}$ can contain states of both types.
The states in $V_\mathrm{n}$ and $V_\mathrm{t}$ are called \emph{nodal}.
A semisplit game must satisfy that the initial state $s_0$ is nodal, and for every non-terminal state $s \in V_\mathrm{n}$, the subgame $\mathcal{G}_s$ contains at least one terminal state.
The second condition ensures that from each nodal state, a terminal state is reachable.
Yet, for an intermediate state, there may be no nodal state in its subgame; then this state is called \emph{dead}.
An edge is now called a \emph{semimove}.
A \emph{submove} is a directed path where nodal states can occur only at the beginning or at the end, and there are only intermediate states in the middle.
Then, a \emph{move} is a submove between two nodal states.

There is a correspondence between a semisplit game and an (ordinary) game.
The \emph{rolled-up} game of a semisplit game is obtained by removing all dead states, and then by replacing each maximal connected component rooted at a non-terminal state with only intermediate states below with one non-terminal state; then all the edges become moves.
A semisplit game $\mathcal{G}'$ is \emph{equivalent} to a game $\mathcal{G}$ if the rolled-up game of $\mathcal{G}'$ is isomorphic to $\mathcal{G}$.
So generally, splitting moves in a game means deriving its equivalent semisplit game.

\begin{figure*}[tb!]\centering
\includegraphics[height=125px]{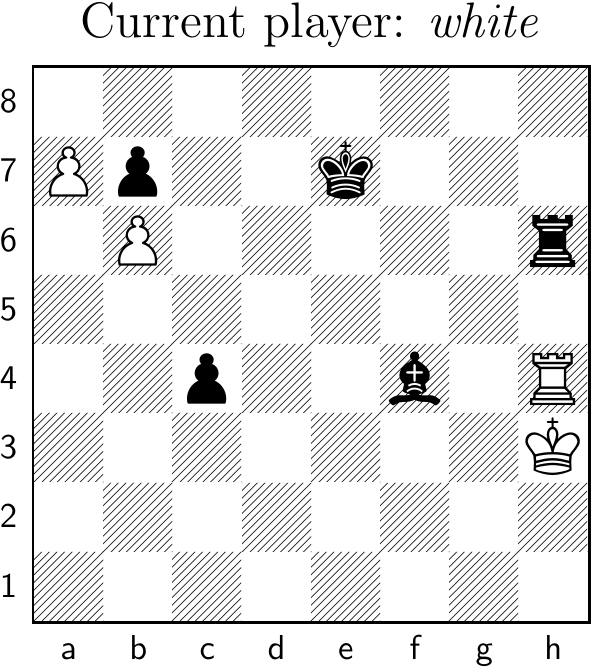}
\hspace{10px}
\includegraphics[height=125px]{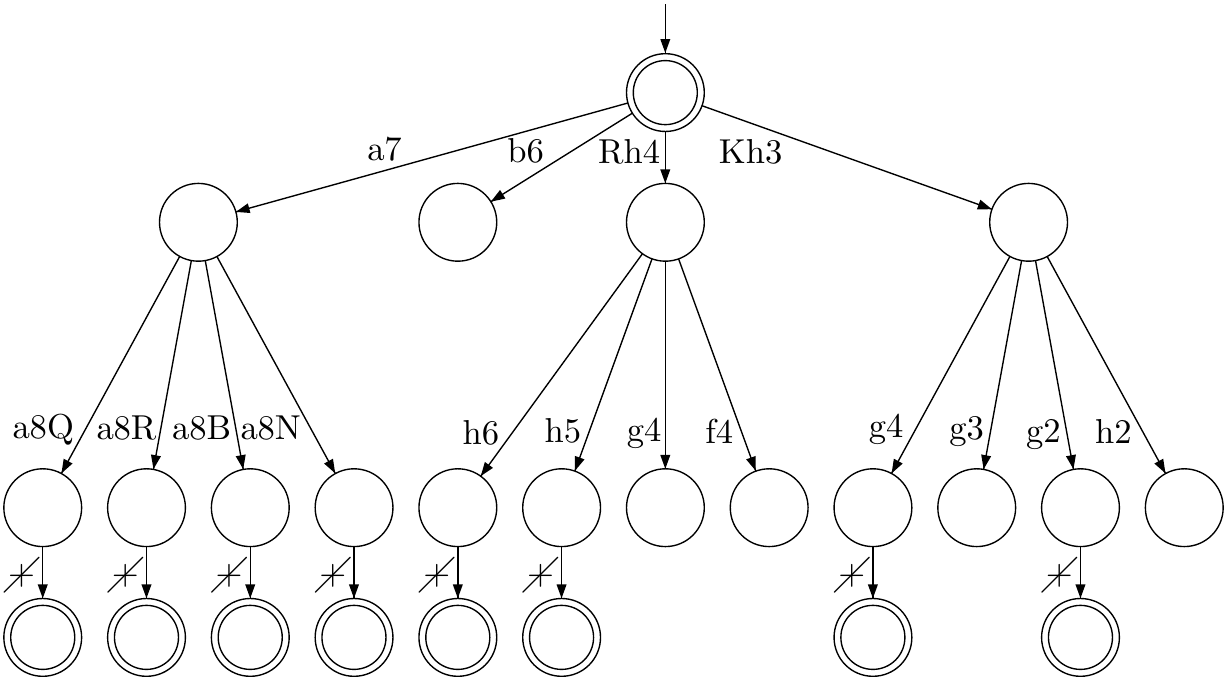}
\hspace{10px}
\includegraphics[height=125px]{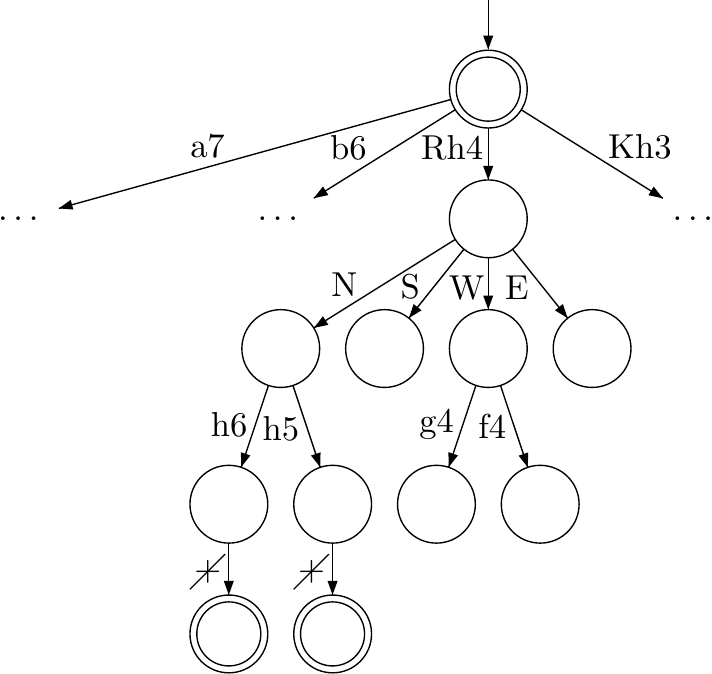}
\caption{A Chess position (left) and the corresponding fragments of two semisplit games of a smaller (middle) and a larger (right) granularity.
There are $8$ legal moves in total ($\text{a7-a8Q}, \ldots, \text{Kh3-g2}$ in the long algebraic notation), which form $8$ edges in the standard game tree.
Nodal states are marked with a double circle; \cancel{+} indicates passed non-check king test.
In the right semisplit game fragment, there are drawn 3 nodal, 9 intermediate, and 5 dead states.}
\label{fig:chessboard_split}
\end{figure*}

An example of two distinct semisplit versions of Chess is shown in Fig.~\ref{fig:chessboard_split}.

\subsection{The Semisplit Algorithm}

We describe a generalization of MCTS that works on an underlying semisplit game.
To distinguish from the standard MCTS algorithm that operates on an ordinary game, the latter is called \emph{orthodox MCTS}.
In the following description, we use standard terminology \cite{Browne2012ASurvey} and focus on the differences with orthodox MCTS.

The simulation phase is shown in Alg.~\ref{alg:semisplit-mcts}, lines~1--13.
Drawing a move at random is realized through backtracking ($\Call{SemisplitRandomMove}{}$).
Given a game state, we choose and apply semimoves in the same way as moves in orthodox MCTS, but we keep the list of legal semimoves computed at each level.
When it happens that the current intermediate state does not have a legal semimove (it is dead), we backtrack and try another semimove.
Thus, a move is always found if it exists, and every legal move has a positive chance to be chosen, although the probability distribution (on legal moves) may be not uniform (in the worst case, the probability of choosing a move can be exponentially smaller in the number of introduced intermediate states).
A single simulation ($\Call{SemisplitSimulation}{}$) just uses the modified random move selection.
Note that this function can fail (line~4), which happens if and only if called for a dead state.

\begin{algorithm}[tb!]\footnotesize
\begin{algorithmic}[1]
\Require{$s$ -- game state}
\Function{SemisplitSimulation}{$s$}
\While{$\mathbf{not} \: s.\Call{IsTerminal}{ }$}
\State $m \gets \Call{SemisplitRandomMove}{s}$
\LineIf{$m = \textbf{None}$}{\Return $\textbf{None}$} \Comment{$s$ is dead}
\State $s \gets s.\Call{Apply}{m}$
\EndWhile
\State \Return $s.\Call{Scores}{ }$
\EndFunction
\Function{SemisplitRandomMove}{$s$}
\ForAll{$a \in \Call{Shuffle}{s.\Call{GetAllSemimoves}{ }}$}
\State $s' \gets s.\Call{apply}{a}$
\LineIf{$s'.\Call{IsNodal}{ }$}{\Return $a$}
\State $m \gets \Call{SemisplitRandomMove}{s'}$
\LineIf{$m \neq \mathbf{None}$}{\Return \Call{Concatenate}{$a,m$}}
\EndFor
\State \Return $\mathbf{None}$ \Comment{No legal move}
\EndFunction
\Function{MCTSIteration}{ }
\State $v \gets \Call{TreeRoot}{ }$
\While{$\mathbf{not} \: v.\Call{State}{ }.\Call{IsTerminal}{ }$}
\If{$v.\Call{FullyExpanded}{ }$}
\State $v \gets v.\Call{UCT}{ }$ 
\Else
\State $a_1 \gets \Call{RandomElement}{v.\Call{UntriedSemimoves}{ }}$
\State $(v',\mathit{scores}) \gets \Call{Expand}{v,a_1}$
\State \algorithmicif\ {$\mathit{scores} = \mathbf{None}$}\ \algorithmicthen\ \textbf{continue}
\State $\Call{Backpropagation}{v', \mathit{scores}}$
\State \Return
\EndIf
\EndWhile
\State $\Call{Backpropagation}{v, v.\Call{State}{ }.\Call{Scores}{ }}$
\EndFunction
\Require{$v$ -- leaf node; $a_1$ -- selected untried semimove}
\Function{Expand}{$v,a_1$}
\State $s \gets v.\Call{State}{ }.\Call{Apply}{a_1}$
\State $\mathit{scores} \gets \Call{SemisplitSimulation}{s}$
\If{$\mathit{scores} = \mathbf{None}$}
\State $v.\Call{RemoveSemimove}{a_1}$
\State \Return $\mathbf{None}$ 
\EndIf
\State $c \gets \Call{CreateNode}{s}$
\State $v.\Call{AddChild}{c,a_1}$
\State \Return $(c, \mathit{scores})$
\EndFunction
\Require{$v$ -- leaf node; $\mathit{scores}$ -- players' final scores}
\Function{Backpropagation}{$v, \mathit{scores}$}
    \While{$v \neq \Call{TreeRoot}{ }$}
        \State $v.\mathit{scoreSum} \gets v.\mathit{scoreSum} + \mathit{scores}[v.\mathit{Player}]$
        \State $v.\mathit{iterations} \gets v.\mathit{iterations} + 1$
        \State $v \gets v.\Call{Parent}{ }$
    \EndWhile
    \State v.$\mathit{iterations} \gets v.\mathit{iterations} + 1$
\EndFunction
\end{algorithmic}
\caption{Vanilla semisplit MCTS.}\label{alg:semisplit-mcts}
\end{algorithm}

The vanilla variant of semisplit MCTS is shown in Alg.~\ref{alg:semisplit-mcts}. 
As orthodox MCTS, it uses the UCT policy in the selection phase \cite{Kocsis2006Bandit}.
But in contrast, semisplit MCTS uses both nodal and intermediate states as tree nodes.
The expansion begins with the selection of an untried semimove (line~20).
Dead states are never added to the MCTS tree.
If the next state turns out to be dead, that semimove is removed from the list in the node, and the search goes back to the MCTS tree, so other untried semimoves are chosen (line~22).
It can happen that all untried semimoves lead to dead states and the node becomes fully expanded; then the search continues according to the UCT policy.

\noindent\textbf{Raw and nodal variants.}
There are two expansion variants, \emph{raw} and \emph{nodal}.
The \emph{raw} variant, given in Alg.~\ref{alg:semisplit-mcts},  adds just one tree node as usual, either intermediate or nodal.
However, this can leave semisplit MCTS behind orthodox MCTS in terms of expansion speed, as a single node in the latter corresponds to a path between nodal states in the former.
In other words, when counting nodal states, orthodox MCTS expands faster.
The \emph{nodal} variant compensates this risk by adding the whole move (path) during a single expansion\ARXIVorAAAI{ (see Appendix, Alg.~2 for pseudocode).}{.}
This may slightly increase the quality of the search when these paths are long and, in particular, cannot be exploited due to slower expansion.
However, the nodal variant may be slightly slower.

\noindent\textbf{Final selection.}
There are several policies to select the final move to play.
A common policy is to choose the one with the best average score \cite{Winands2017MCTSBoardGames} among tried moves.
When it comes to selecting the final move to play, semisplit MCTS greedily chooses the best semimove till the first next nodal state.
If the move goes outside of the MCTS tree constructed so far, the raw variant chooses the remaining semimoves uniformly at random.
This can happen when the branching factor is very large compared to the number of iterations.
In the nodal variant this cannot occur, as all leaves added to the MCTS tree are nodal states.

\noindent\textbf{Combined variants and roll-up.}
Possible variants of semisplit MCTS involve combining both designs and using them selectively in different phases.
There are two natural variants: orthodox design in the MCTS tree phases (selection and expansion) combined with semisplit design in the simulation phase, and the opposite variant.

Another proposed variant is adaptive, applied in the MCTS tree phases.
The idea is to use semisplit design at first, to improve efficiency and deal with potentially large branching factor, and then switch to orthodox design to provide more conservative evaluation.
The \emph{roll-up} variant uses semisplit design in the MCTS tree with the modification as follows\ARXIVorAAAI{ (see Appendix, Alg.~3 for pseudocode).}{.}
Whenever a node $v$ of an intermediate state in the MCTS tree is fully expanded, i.e., all its children $v_1,\ldots,v_k$ were tried, the algorithm removes $v$ and connects $v_1,\ldots,v_k$ directly to its parent $v'$. 
Then, the edges to them are submoves obtained from concatenating the submove from $v'$ to $v$ and the submoves from $v$ to $v_i$.
In this way, semisplit design is limited to the expansion phase, as the algorithm switches to orthodox design in the parts of the MCTS tree that become fully expanded.
The roll-up variant can be used to test the impact of semisplit design in this phase alone.

\noindent\textbf{Action-based heuristics.}
Common general knowledge-free enhancements of MCTS are online learning heuristics, which estimate the values of moves by gathering statistics.
We test the two most fundamental methods, MAST and RAVE \cite{finnsson2008simulation,gelly2011monte,Winands2017MCTSBoardGames}.
MAST globally stores for every move (actually, for every set of moves with the same representation) the average result of all iterations containing this move; it is used in simulations in place of a uniformly random choice.
RAVE stores similar statistics locally in each MCTS node for available moves and uses them to bias the choice in UCT.
For both techniques, there are many policies proposed that differ in details.

\setlength{\tabcolsep}{5pt}
\begin{table*}[!tb]\centering\renewcommand{\arraystretch}{1.0}\small
\newcommand{\col}[1]{\multicolumn{1}{c|}{#1}}
\newcommand{\colFour}[1]{\multicolumn{4}{c|}{#1}}
\newcommand{\tb}[1]{\textbf{#1}}
\newcommand{\nb}[1]{#1}
\caption{Flat Monte-Carlo results (random simulations with gathering scores).
The 2st and 3rd column show the speed measured in the number of computed respective nodal states and simulations per second.
The 4th and 5th columns show the mean simulation depth measured resp. in nodal states and all states (computed dead states are also included).
The last 6th column shows the branching factor calculated as the number computed (semi)moves divided by the number of states.
}\label{tab:benchmark_stats}
\begin{tabular}{|l|r|r|r|r|r|}\hline
Game                                 & Nodal states/sec.  & Sim./sec.  & Mean nodal states/sim. & Mean all states/sim. & Mean br.\ factor (all states)\\\hline
Breakthrough {\scriptsize(orthodox)} & 2,388,827 (100\%) &   37,272 &                       64 &                 64 & 25.69  \\
Breakthrough {\scriptsize(@Mod)}     & 5,717,972 (239\%) &   78,120 &                       73 &                157 & 7.71   \\
Breakthrough {\scriptsize(@ModShift)}& 7,558,711 (316\%) &  220,074 &                       34 &                164 & 3.12   \\
Chess {\scriptsize(orthodox)}        &   308,157 (100\%) &      976 &                      316 &                316 & 22.83  \\
Chess {\scriptsize(@Mod)}            & 1,556,036 (505\%) &    5,971 &                      261 &              1,494 & 2.96   \\
Chess {\scriptsize(@ModPlus)}        & 1,390,428 (451\%) &    5,341 &                      260 &              2,194 & 2.31   \\\hline
\end{tabular}
\end{table*}

MAST and RAVE can be adapted to semisplit in a straightforward way if semisplit design is used both in the MCTS tree and simulation.
In combined variants, they require some adjustments, which rely on dividing (sub)moves into semimoves or merging semimoves into moves.

\noindent$\bullet$\ \emph{\mastsplit} and \emph{\ravesplit}:
The basic modification of MAST adapted to semisplit simply stores separate statistics for each semimove.
Then, when evaluating the score of a longer submove, we have to somehow combine the result from the scores of the included semimoves.
Here, we propose the arithmetic mean of the scores of the semimoves. However, if a semimove has not been tried, then the maximum reward is returned as the final score of the whole submove.
\mastsplit can be applied to any variant of MCTS.

\emph{\ravesplit} works analogously, i.e., it splits every submove into single semimoves.
Statistics for a semimove in a tree node are updated if this semimove was used explicitly at the node or later in the iteration, either directly or possibly as a part of a submove.
\ravesplit is easily enabled only in the combinations using semisplit design in the selection and expansion phases, i.e. when the domain of semimoves in these phases correspond to the stored statistics.

Depending on the implementation, split variants can be much faster than regular heuristics due to a much smaller domain of semimoves (i.e., we can use faster data structures for storing statistics).
However, obtained samples are less reliable, as semimoves carry less information than full moves.

\noindent$\bullet$\ \emph{\mastjoin} and \emph{\ravejoin}:
These variants merge semimoves into full moves.
\mastjoin and \ravejoin store statistics only for whole moves.
Of course, they are available only in the cases where we evaluate only moves in the corresponding MCTS phases.

\noindent$\bullet$\ \emph{\mastcontext} and \emph{\ravecontext}:
To partially overcome the weakness of less reliable samples, we also propose \emph{context} variants.
They lead to storing sample values closer to those used in orthodox design.
The main idea of \mastcontext is entwined with \emph{N-gram-Average Sampling Technique} (\emph{NST}) \cite{tak2012ngrams}, which gathers statistics for fixed-size sequences of moves.
\mastcontext maintains statistics for submoves of different lengths.
When the iteration is over, for each move applied in the iteration, the statistics are updated not only for that move but also for each of its prefixes.
The \emph{context} of a state is the submove from the last preceding nodal state to this state.
Thus, a context is a submove that is a prefix of some move.
While selecting the best submove in the simulation phase, we use the statistics of the submoves concatenated to the current context.
In \ravecontext, MCTS nodes store statistics for each child just as in the regular RAVE.
The difference is in updating them; the statistic of a semimove is updated only if the iteration contains the same semimove played in the same context.
The context variants are available for every variant of MCTS (in particular, for roll-up) because they store statistics for all submoves that may be ever needed.

\noindent$\bullet$\ \emph{\mastmix} and \emph{\ravemix}:
For most MCTS variants, we have more than one choice for variants of action-based heuristics.
This leads to the possibility of using more than one simultaneously.
The variants of the heuristics differ in speed of gathering samples but also in their quality, e.g., \mastsplit gathers samples faster than \mastcontext, but they are less reliable.
Hence, \mastmix combines \mastsplit with \mastcontext, and \ravemix combines \ravesplit with \ravecontext.
All statistics are updated separately according to both split and context strategies.
For both heuristics, we have an additional parameter -- the \emph{mix-threshold}.
During the evaluation, when the number (weight) of samples in the context heuristic is smaller than the mix-threshold, the score is evaluated according to the split variant; otherwise, the context statistics are used.

\ARXIVorAAAI{An extended discussion on these heuristics and their available combinations can be found in Appendix, Sec.~1.}{}

\section{Implementation and Experiments}

Our test set consists of 12 board games, well known in GGP (Amazons, Breakthrough, Breakthru, Chess, Chess without check, English Draughts, Fox And Hounds, Go, Knightthrough, Pentago, Skirmish, and The Mill Game).
\ARXIVorAAAI{Appendix, Sec.~3 contains the precise rules of the used versions.}{}

\noindent\textbf{Split Strategies.}
For a given game, there are many equivalent split versions.
Usually, they are derived through a manual implementation of the algorithm computing legal moves and states.
For the purposes of experiments, we use abstract game descriptions in the Regular Boardgames and derive semisplit games through \emph{split strategies}, which are algorithms taking as the input a pure definition of the game rules (without any heuristic information for good playing).
Thus, they can be considered knowledge-free and derive splittings automatically in a systematic way across different games.

Although split strategies might be interesting by themselves, they are developed here for the purposes of experiments.
This part is to create an example environment for testing semisplit MCTS, and thus it is not the focus of the current paper.
We note that using such split strategies would not be possible effectively without semisplit MCTS, as they introduce dead states.
Similar effects could be obtained via other approaches; in particular, one could manually implement each semisplit game, apart from any GGP system.

\ARXIVorAAAI{The omitted technical details of RBG language, split strategies, and how they affect each particular game are given in Appendix~Sec.~2 and~3.}{}
Below, we describe the intuitive meaning of split strategies.
Each split strategy is defined via a subset of three components.

\noindent$\bullet$\ \emph{Mod}:
This is a basic component of relatively low granularity.
Every semimove corresponds to an action modifying either a single square on the board or a variable; thus there is introduced a semimove for each elementary modification of the game state.
A move in a chess-like game is split into two semimoves: one for selecting and grabbing a piece from the board and one for dropping it on the destination square.
For modifying more squares, more semimoves are introduced accordingly (e.g., Amazons).
Also, it separates the final decision in a move, e.g., in Chess, the king check is performed in a final semimove; in Go, the player first chooses whether to pass or to put a stone. 
An example of the resulting semisplit game tree is shown in Fig.~\ref{fig:chessboard_split}(middle).

\noindent$\bullet$\ \emph{Plus}:
This component introduces a semimove for every single decision (branch) specified by the rules except those made iteratively.
It commonly involves choosing the direction of the movement (Amazons, Chess, English Draughts) and move type (capture or two movements in Breakthru).
It does not split decisions that are repeated, such as stopping on a square while moving in a chosen direction (e.g., rook upwards move).
An example of the resulting semisplit game tree is shown in Fig.~\ref{fig:chessboard_split}(right).

\noindent$\bullet$\ \emph{Shift}:
This component splits square selection when it consists of more than one decision.
It commonly separates the selection of a column and a row of the board (all games) and also divides movements of some pieces (e.g., knight -- first long and then short hop).

By combining these components, we can build split strategies such as $\emph{Mod}$, $\emph{ModShift}$, or $\emph{ModPlusShift}$.
Additionally, the algorithm optimizes the semisplit game by removing some indecisive splits (i.e., semimoves that are always the only available choice) when possible.

\subsection{Results}

\setlength{\tabcolsep}{3pt}
\begin{table}[!tb]\centering\small\renewcommand{\arraystretch}{0.05}
\newcommand{\colTwo}[1]{\multicolumn{2}{c|}{#1}}
\newcommand{\colBTwo}[1]{\multicolumn{2}{|c|}{#1}}
\newcommand{\wid}[0]{152px}
\newcommand{\colG}[1]{\includegraphics[angle=0,width=\wid]{figures_winrates/#1.pdf}}
\newcommand{\notbot}[1]{ \vbox{\hbox{\strut #1}{\tiny \hbox{  \vphantom{-}}} }}
\newcommand{\notbotX}[1]{\tiny \vbox{\hbox{\strut #1}{\tiny \hbox{  \vphantom{.}}} }}
\caption{Win rates of semisplit agents.
The bars show the grand average in timed (upper, red) and fixed (lower, blue) settings. Timed results are also given separately for each game; green, yellow, and red indicate statistically significant \emph{better}, \emph{same}, and \emph{worse} performance, respectively.
A game is counted for better performance if the whole 95\% confidence interval is above 50\%; worse is symmetrical.
}\label{tab:agents_averages}
\begin{tabular}{|l|c|}\hline
\multicolumn{1}{|c|}{Agent}   &  Win rates vs. \orthodox\phantom{\large X} \\[3pt]\hline
\notbot{\semisplitV{S-raw}{S}{Mod}}    & \colG{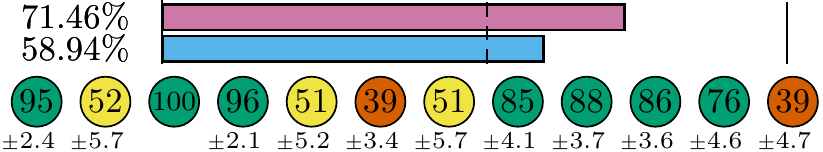} \\\hline
\notbot{\semisplitV{S-nodal}{S}{Mod}} & \colG{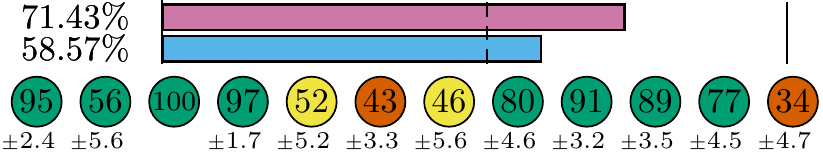} \\\hline
\notbot{\semisplitV{S-nodal}{O}{Mod}} & \colG{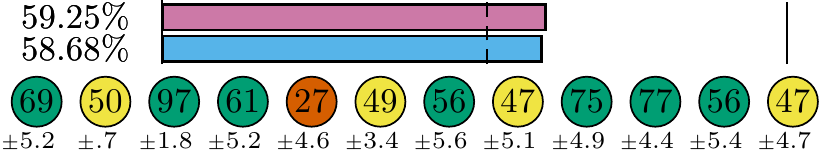} \\\hline
\notbot{\semisplitV{O}{S}{Mod}}       & \colG{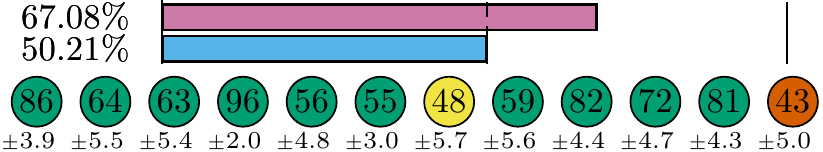} \\\hline
\notbot{\semisplitV{R-nodal}{S}{Mod}} & \colG{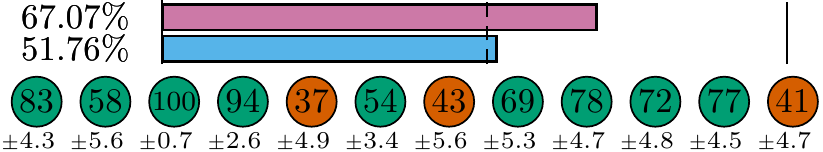} \\\hline
\notbot{\semisplitV{R-nodal}{O}{Mod}} & \colG{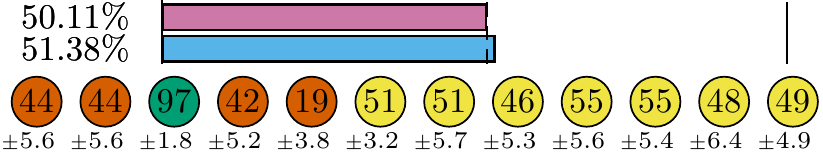} \\\hline
&  \includegraphics[angle=90,width=\wid]{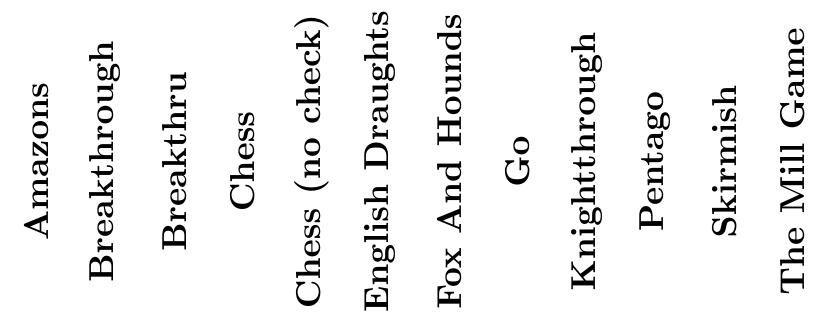} \\\hline
\notbot{\semisplitV{S-nodal}{S}{ModPlus}} & \colG{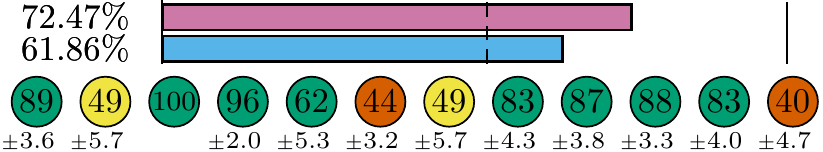} \\\hline
\notbot{\semisplitV{S-raw}{S}{ModPlus}} & \colG{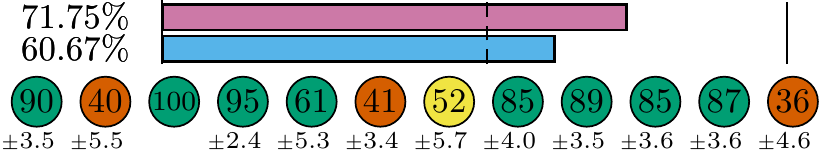} \\\hline
\notbot{\semisplitV{S-nodal}{S}{ModShift}} & \colG{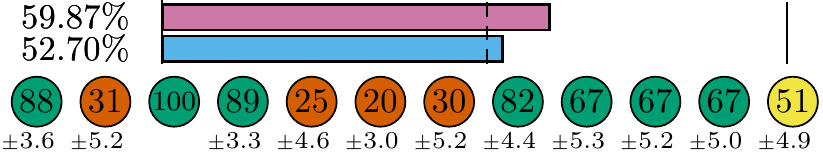} \\\hline
\notbot{\semisplitV{S-nodal}{S}{ModPlusShift}} & \colG{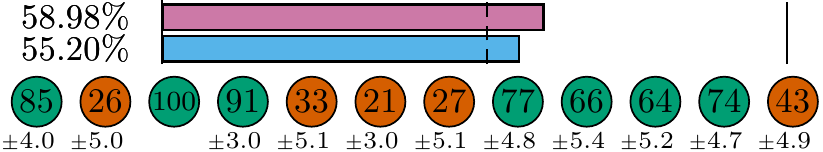} \\\hline
\notbot{\semisplitV{O}{S}{ModPlus}} & \colG{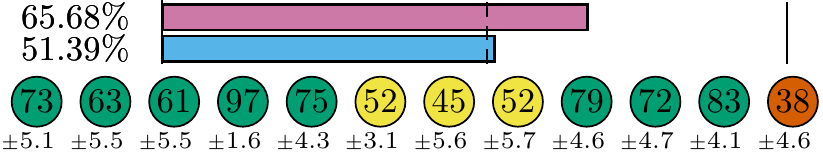} \\\hline
\multicolumn{1}{|c|}{Agent}     &  Winrates vs. \orthodoxMR\phantom{\LARGE X} \\[4pt]\hline
\notbot{\semisplitV{S-nodal,{\tiny\ravesplit}}{S,{\tiny\mastsplit}}{Mod}} & \colG{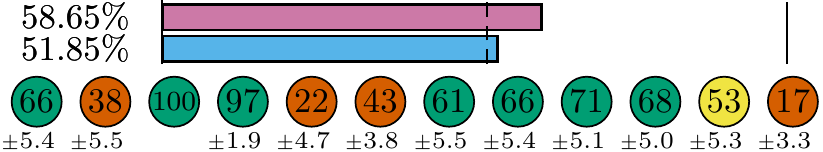} \\\hline
\notbot{\semisplitV{S-nodal}{S,{\tiny\mastsplit}}{Mod}}    & \colG{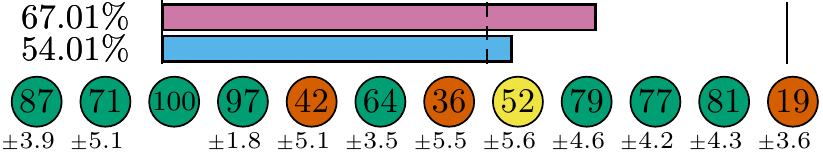} \\ \hline
\notbot{\semisplitV{S-nodal}{S,{\tiny\mastmix}}{Mod}}    & \colG{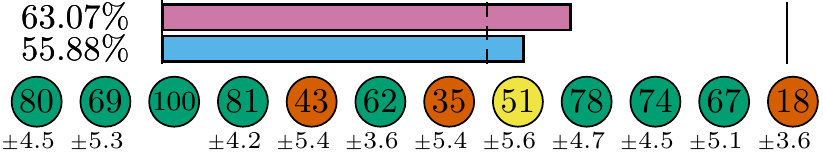} \\\hline
\notbot{\semisplitV{O,\ravejoin}{S,\mastsplit}{Mod}}  & \colG{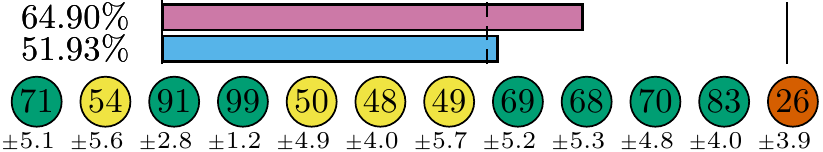} \\ \hline
\notbot{\semisplitV{R-nodal}{S,{\tiny\mastsplit}}{Mod}}    & \colG{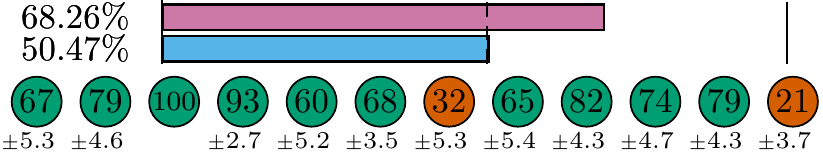} \\\hline
\end{tabular}
\end{table}

\begin{figure*}[htb!]\centering
\includegraphics[width=\textwidth]{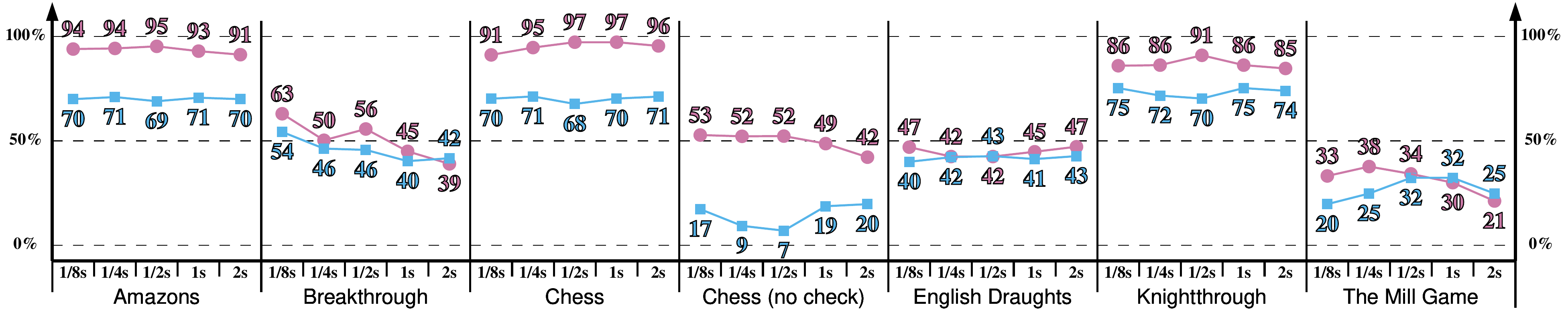}
\caption{The results of \semisplit{S-nodal}{S}{Mod} versus $\mathbb{O}$ for different time limits (circle) and for equivalent states budgets (square).}
\label{fig:experiments-chart}
\end{figure*}

\noindent\textbf{Setup and notation.}
\ARXIVorAAAI{The full technical setup and methodological details are given in Appendix, Sec.~4.}{}
All parameters (e.g., the exploration factor, MAST and RAVE policies) were set according to the recommendations in the literature \cite{Finnsson2010Learning,sironi2016comparison} and keep the same for every agent.
Semisplit agents were tested against the orthodox ones: both the vanilla MCTS agent (denoted by \orthodox) and the enhanced MCTS agent (\orthodoxMR) were used as baselines.
Our semisplit agents are denoted by $\mathbb{S}$ with indices describing the variant: ``S'' means semisplit design and ``O'' means orthodox design, which can be independently used in the MCTS tree or in simulations; ``R'' is the roll-up variant.
Raw and nodal variants and which variants of MAST and RAVE are used, if applied, are also indicated.
Finally, the used split strategy is denoted after~``@''.

We tested agents in two settings: the \emph{timed setting}, where agents have the same amount of time (0.5s) per turn, and the \emph{fixed setting}, where agents have a limited number of nodal states to compute.
These settings are correlated by letting the states limit equal to the number of states computed by the baseline orthodox agent performing within the given time limit.
Hence, the fixed setting is used to observe the impact of semisplit design apart from efficiency benefits.
Note that this is different than the common measurement with a fixed number of simulations and gives a better correspondence with the timed setting, as splitting alters simulation~length.

\noindent\textbf{Statistics.}
Tab.~\ref{tab:benchmark_stats} shows illustratively how simulation statistics change under semisplit design.
We observe a significant speed-up in terms of the number of traversed states, and thus of simulations in a given time limit.
Obviously, semisplit design increases the depth of a simulation and reduces the mean branching factor.
When the branching factor is already small, introducing more splits slows down.
The real threshold strongly depends on the game and the implementation.
Note that the mean depth of a simulation measured in nodal states is also altered, which can independently influence the number of simulations.

\noindent\textbf{Comparison of agents.}
Table~\ref{tab:agents_averages} contains three parts.
The first part shows the results of variants of semisplit MCTS without action-based heuristics and using the Mod split strategy (one of the least granularity).
All variants get better results than orthodox MCTS, except for \semisplit{R-nodal}{O}{Mod}.
There is no visible difference between the raw and nodal variants.
The efficiency benefits alone are most significant when the agent uses semisplit in simulations.
We also see an advantage in the fixed setting when semisplit is applied in the MCTS tree.
While \semisplit{S-raw}{S}{Mod} gets the largest mean, \semisplit{O}{S}{Mod} is the most robust, winning in ten games and losing only in one.
The particular case of \semisplit{R-nodal}{O}{Mod} shows the effect of semisplit applied only to the expansion phase, which results in a kind of an unprunning method \cite{chaslot2008progressive} -- this is beneficial only for Breakthru, which has a very large branching factor.

The second part shows four representative combinations with split strategies of a larger granularity.
Here, the nodal variant has a slight advantage over raw.
The results follow the same pattern, yet ModPlus seems to be slightly better than Mod, but ModShift already worsens the results.

The third part shows selected combinations of agents equipped with RAVE and MAST against \orthodoxMR.
\semisplit{S-nodal,\ravesplit}{S,\mastsplit}{Mod} is the simplest variant; it gives positive results but is not that strong as previously.
Surprisingly, RAVE does not perform that well here; thus it is better to omit it.
\semisplit{S-nodal}{S,\mastmix}{Mod} gives the best results in the fixed setting, but the computation cost of \mastmix does not overcome the benefits.
The best and most robust agent among tested ones is the roll-up variant with semisplit simulations and \mastsplit.

Fig.~\ref{fig:experiments-chart} shows how the win rate changes for different time limits and in relation to the fixed setting.
In most cases, the results are kept consistent for different limits, especially in the games where semisplit is clearly beneficial.

\section{Conclusions}

We have introduced a family of Monte-Carlo Tree Search variants that work on semimoves -- arbitrarily split game moves.
The algorithm is based on the idea of lazy move computation, yet it has many possible variants.
We applied split design for single-action games for the first time and revealed the impact on agents' results on a wider set of games (as previously, only Amazons was tested).
Furthermore, we tested different configurations concerning the selective application in MCTS phases, split granularity, and variants of action-based heuristics.
The developed framework allows testing the effects of (semi)splitting for more than 600 combined variants of semisplit, action-based heuristics, and split strategies, for any game described in RBG.

The impact of using semisplit is generally beneficial.
First, it greatly improves search efficiency (typically, 3--5 times more nodal states/sec.).
Moreover, for many games, the playing strength is improved over its orthodox counterpart, also when both algorithms are capped to the same performance.
Even with general and blind split strategies, we were able to obtain win rates larger than 70\% on about half of the test set (e.g., \semisplit{S-nodal}{S}{Mod} or \semisplit{R-nodal}{S,\mastsplit}{Mod}).
For comparison, the benefits are larger than those from adding action-based heuristics to the vanilla orthodox agent (the mean win rate of \orthodoxMR vs. \orthodox\ on our test set is 53\% in timed and 65\% in fixed setting).

More detailed results show that using semisplit in simulations gives a large efficiency boost and generally is not harmful in terms of their quality.
This may be useful also apart from game playing, as simulations are applied to many single-player problems \cite{wang2020tackling}.
On the other hand, using semisplit in the MCTS tree gives some benefits in the quality of iterations, yet it is riskier, as sometimes it is consistently harmful (e.g., The Mill Game).
However, on average, almost none of the tested variants is worse in the fixed setting than the orthodox baseline.

The conducted research can be seen as pioneering, as there are many directions for future research, e.g.:\\
$\bullet$ There should be developed methods for choosing the most suitable semisplit variant and split strategy for a given game.\\
$\bullet$ The used parameters were the same for both agent types and were tuned rather for orthodox agents according to the literature. This indicates that after a tuning, semisplit should achieve even better results.\\
$\bullet$ We focused on the practical aspect, yet there are interesting theoretical questions as to how strongly splitting the game can distort the agent's results, and how hard is the problem? to determine whether splitting will be beneficial.\\
$\bullet$ Semisplit can be combined with prior knowledge \cite{gelly2007combining} or neural networks. 
It can reduce action space and improve efficiency, especially when simulations are combined with neural networks \cite{cotarelo2021improving}.

\section{Acknowledgements}

This work was supported in part by the National Science Centre, Poland under project number 2021/41/B/ST6/03691 (Jakub Kowalski, Marek Szyku{\l}a).

\bibliography{bibliography}

\newpage\onecolumn
\section{\LARGE Appendix}

\setcounter{secnumdepth}{2}

\setcounter{algorithm}{1}
\setcounter{table}{2}
\setcounter{figure}{1}
\section{Details of Semisplit MCTS}

Alg.~\ref{alg:semisplit-nodal} shows the expansion function in the nodal variant, which is called in Alg.~1~line~21 in the main text.

Alg.~\ref{alg:rollup-mcts} shows the roll-up function, which checks and removes a given node of an intermediate (non-nodal) state.
It is called during the backpropagation phase for all MCTS nodes involved in the iteration, in bottom-up order (Alg.~\ref{alg:backpropagation} line~15).

\begin{algorithm}[htb!]\footnotesize
\begin{algorithmic}[1]
\Require{$v$ -- leaf node in MCTS tree}
\Require{$a_1$ -- selected untried semimove}
\Function{ExpandNodal}{$v,a_1$}
\State $s \gets v.\Call{State}{ }$
\State $m \gets \Call{SemisplitRandomMove}{s.\Call{Apply}{a_1}}$
\If{$m = \mathbf{None}$}
\State $v.\Call{RemoveSemimove}{a_1}$
\State \Return $\mathbf{None}$
\EndIf
\State $m' \gets \Call{Concatenate}{a_1,m}$
\ForAll{$a \in m'$} \Comment{{\small For all semimoves $a$ in $m'$ in the order}}
\State $s \gets s.\Call{Apply}{a}$
\State $c \gets \Call{CreateNode}{s}$
\State $v.\Call{AddNode}{c, a}$
\State $v \gets c$
\EndFor
\State $\mathit{scores} \gets \Call{SemisplitSimulation}{v.\Call{State}{ }}$
\State \Return $(v, \mathit{scores})$ 
\EndFunction
\end{algorithmic}
\caption{The nodal expansion variant.}\label{alg:semisplit-nodal}
\end{algorithm}

\begin{algorithm}[htb!]\footnotesize
\begin{algorithmic}[1]
\Require{$v$ -- node (of intermediate state) to roll-up}
\Function{RollUp}{$v$}
    \If{$v.\Call{FullyExpanded}{ }$}
        \State $\mathit{childrenCount} \gets v.\Call{Children}{ }.\Call{Length}{ }$
        \If{$v.\mathit{iterationCount} \geq \mathit{childrenCount}$}
            \State $v' \gets v.\Call{Parent}{ }$
            \State $m \gets v'.\Call{Submove}{v}$ \Comment{Get the submove from $v'$ to $v$}
            \State $v'.\Call{RemoveChild}{v}$
            \ForAll{$(c,a) \in v.\Call{Children}{ }$}
                \State $m' \gets \Call{Concatenate}{m, a}$
                \State $v'.\Call{AddChild}{c, m'}$
                \State $v'.\mathit{AMAF}[m'] \gets c.\mathit{AMAF}[a]$ \Comment{If RAVE is used}
            \EndFor
        \EndIf
    \EndIf
\EndFunction
\end{algorithmic}
\caption{The roll-up procedure.}\label{alg:rollup-mcts}
\end{algorithm}

\subsection{MAST and RAVE}

Commonly used general knowledge-free enhancements of MCTS are online-learning heuristics, which estimate the values of moves by gathering statistics.
Two general methods are Move-Average Sampling Technique (MAST) and Rapid Action Value Estimation (RAVE) \cite{finnsson2008simulation,gelly2011monte,Winands2017MCTSBoardGames}.
For both techniques, there are many policies proposed that differ in detail.

MAST globally stores for every move the average result of all iterations (samples) involving this move.
Those values are updated after each iteration, and the score of each move is updated as many times as the move was applied in the iteration.
The statistics are stored for each player separately.
They are later used in the simulation and (optionally) expansion phases in place of the random selection.
For each move from the set of all legal moves at some state, MAST defines the probability that this move will be selected.
This probability distribution is defined depending on a particular policy, e.g., Gibb's distribution \cite{finnsson2008simulation}, Roulette wheel \cite{powley2013bandits} or $\epsilon$-greedy \cite{tak2012ngrams}.
According to the previous research, $\epsilon$-greedy is universally considered to be the best policy \cite{powley2013bandits,tak2012ngrams}.
It is also common to assume that if the move was not tried any time before, then its score is the maximum available reward, so it will be preferred to play as soon as possible.

An improvement for MAST is the \emph{decaying} technique \cite{tak2014decaying}.
It comes from the insight that certain moves that are good globally in the game may not be so strong in latter phases.
In other words, that the global average may be not equally adequate for all stages (subgames) of the game.
Decaying means that the weights of all the collected samples are multiplied by the \emph{decay factor}, hence new samples gathered will have more impact on the average score.
A few methods of decaying are known and proposed in \cite{tak2014decaying}: \emph{move decay} takes place after the effective move is applied in the game, \emph{batch decay} applies decaying after a fixed number of simulations, and \emph{simulation decay} applies decaying after each played move (of either the agent or the opponent).
The last option was considered to be possibly the best.

Monte Carlo Tree Search needs many iterations to gather enough samples to differentiate the most promising moves.
At the beginning, many choices are made randomly.
RAVE tries to shorten this phase by storing additional statistics in nodes.
Similarly to MAST, it stores the average reward for moves, but locally in MCTS nodes.
These statistics are called \emph{All Moves As First} (\emph{AMAF}).
For each node, each child with move $m$ stores the average score (for the player of the node) from all iterations that passed through the node and that include $m$ played at the node or at any place below.
This is in contrast with the regular statistic of $m$ in UCT, which is gathered only from iterations that play $m$ at the current node.
In this way, AMAF collects much more samples, which are less reliable, however.

The AMAF statistics are used to bias the choice in the selection phase by modifying the UCB formula.
The priority for each node is calculated as the weighted mean of the regular average score and the average stored for all simulations.
There are many different formulas to calculate such weights
\cite{cazenave2015grave,finnsson2008simulation,gelly2011monte,sironi2016comparison}.
The general rule is that for fewer iterations the statistics gathered by RAVE are more relevant due to a larger number of samples than the MCTS score, but then they lose significance as they do not distinguish between the situations where moves are played.

\begin{algorithm}[htb!]\footnotesize
\begin{algorithmic}[1]
\Require{$\mathit{semimoves}$ -- list of semimoves to choose from}
\Require{$\mathit{player}$ -- current player}
\Function{ChooseSemimoveWithMAST}{$\mathit{semimoves}, \mathit{player}$}
    \If{$\Call{RandomRealNumber}{0,1} \geq \epsilon$}
        \State $\mathit{bestSemimoves} = \argmax_{m \in \mathit{semimoves}}(\mathit{MASTstatistics}[\mathit{player}].\Call{Score}{m})$
        \State \Return $\Call{RandomElement}{\mathit{bestSemimoves}}$
    \Else
        \State \Return $\Call{RandomElement}{\mathit{semimoves}}$
    \EndIf
\EndFunction
\smallskip
\Require{$s$ -- game state}
\Function{SemisplitMASTMove}{$s$}
\State $semimoves \gets s.\Call{GetAllSemimoves}{ }$
\While{$\mathbf{not} \: \mathit{semimoves}.\Call{Empty}{ }$}
    \State $a \gets \Call{ChooseSemimoveWithMAST}{\mathit{semimoves}, s.\Call{Player}{ }}$
    \State $s' \gets s.\Call{Apply}{a}$
    \LineIf{$s.\Call{IsNodal}{ }$}{\Return $a$}
    \State $m \gets \Call{SemisplitMASTMove}{s'}$
    \LineIf{$m \neq \mathbf{None}$}{\Return $\Call{Concatenate}{a,m}$}
\EndWhile
\State \Return $\mathbf{None}$
\EndFunction
\end{algorithmic}
\caption{MAST working on semimoves.}\label{alg:MAST}
\end{algorithm}

Both MAST and RAVE can be adapted to semisplit game trees and work in the pure semisplit design in a straightforward way, i.e., in the same manner as in the orthodox design.
Alg.~\ref{alg:MAST} shows the modified move selection in the simulation phase, where a move is selected with the $\epsilon$-greedy policy.
We use $\Call{SemisplitMASTMove}{}$ in place of $\Call{SemimoveRandomMove}{}$ that is called in Alg.~1, line~3.
If statistics are used to select move in expansion phase, then line 20 of Alg.~1 is also replaced with $a_1 \gets $ \Call{ChooseSemimoveWithMAST}{$v.\Call{UntriedSemimoves}{ }, v.\Call{Player}{ }$}.
RAVE just modifies the UCT selection (Alg.~1, line~18) and backpropagation (Alg.~\ref{alg:backpropagation}), and includes transferring AMAF values in the case of the roll-up variant (Alg.~\ref{alg:rollup-mcts}, line~11).

\begin{algorithm}[htb!]\footnotesize
\newlength{\groupmarginC}
\settowidth{\groupmarginC}{{\scriptsize{MAST\ }\hspace{-1.2cm}}$\begin{cases}\end{cases}$}
\newlength{\groupmarginD}
\settowidth{\groupmarginD}{{\scriptsize{RAVE\ }\hspace{-1.2cm}}$\begin{cases}\end{cases}$}
\newlength{\groupmarginE}
\settowidth{\groupmarginE}{{\scriptsize{Roll-up\ }\hspace{-1.2cm}}$\begin{cases}\end{cases}$}
\begin{algorithmic}[1]
\Require{$v$ -- last node of iteration in MCTS tree}
\Require{$\mathit{scores}$ -- final scores for each player of iteration}
\Require{$\mathit{appliedSubmoves}$ -- list of submoves applied in simulation phase}
\Function{Backpropagation}{$v, \mathit{scores}, \mathit{appliedSubmoves}$}
    \State $\mathit{scores} \gets \mathit{state}.\Call{Scores}{ }$
    \ForAll{$(m, \mathit{player}) \in \mathit{appliedSubmoves}$}  \Comment{with MAST}
        \State $\mathit{MASTstatistics}[\mathit{player}].\Call{Update}{m,\mathit{scores}}$  \Comment{with MAST}
    \EndFor
    \While{$v \neq \Call{TreeRoot}{ }$}
        \State $v.\mathit{scoreSum} \gets v.\mathit{scoreSum} + \mathit{scores}[v.\mathit{Player}]$
        \State $v.\mathit{iterations} \gets v.\mathit{iterations} + 1$
        \State $v' \gets v.\Call{Parent}{ }$
        \State $m \gets v'.\Call{Submove}{v}$ \Comment{with MAST or RAVE}
        \smallskip
        \State $\mathit{MASTstatistics}.\Call{Update}{m, \mathit{scores}, v'.\Call{Player}{ }}$ \Comment{with MAST}
        \smallskip
        \State $\mathit{appliedSubmoves}[v'.\Call{Player}{ }].\Call{Append}{m}$ \Comment{with RAVE}
            \ForAll{$(c, a) \in v.\Call{Children}{ }$} \Comment{with RAVE}
                \If{$a \in \mathit{appliedSubmoves[v'.\Call{Player}{ }]}$} \Comment{with RAVE}
                    \State $v'.\mathit{AMAF}[a].\Call{Update}{\mathit{score}[v'.\Call{Player}{ }]}$ \Comment{with RAVE}
                \EndIf
            \EndFor
        \smallskip
        \State \Call{RollUp}{$v$} \Comment{with roll-up}
        \smallskip
        \State $v \gets v'$
    \EndWhile
    \State $\Call{TreeRoot}{ }.\mathit{iterations} \gets \Call{TreeRoot}{ }.\mathit{iterations} + 1$
\EndFunction
\end{algorithmic}
\caption{Variants of backpropagation, depending on whether MAST, RAVE, or roll-up is used, respectively.}\label{alg:backpropagation}
\end{algorithm}

\subsection{Available Combinations}

\begin{table}[!htb]\footnotesize\renewcommand{\arraystretch}{1.2}
\caption{Available combinations of the variants of MCTS and action-based heuristics.}\label{tab:heur_combinations}
\setlength{\tabcolsep}{6pt}
\newcommand{\checkss}[1]{\checkmark\textsuperscript{#1}}
\newcommand{\ccheckss}[1]{\bcancel{\checkmark\textsuperscript{#1}}}
\begin{center}
\begin{tabular}{|l|l|c|c|c|c|c|c|c|c|}\hline
\multicolumn{2}{|c|}{MCTS variant} & \multicolumn{2}{c|}{Standard} & \multicolumn{2}{c|}{Split} & \multicolumn{2}{c|}{Join} & \multicolumn{2}{c|}{Context} \\\hline
Tree      & Sim.      & MAST        & RAVE        & MAST        & RAVE        & MAST        & RAVE        & MAST        & RAVE \\
\hline
orthodox  & orthodox  & \checkss{J} & \checkss{J} & \checkmark  &         --  & \checkmark  & \checkmark  & \ccheckss{J}& \ccheckss{J}\\
orthodox  & semisplit &         --  &         --  & \checkmark  &         --  &         --  & \checkmark  & \checkmark  & \ccheckss{J}\\
semisplit & orthodox  &         --  &         --  & \checkmark  & \checkmark  & \checkss{*} &         --  & \checkmark  & \checkmark \\
semisplit & semisplit & \checkss{S} & \checkss{S} & \checkmark  & \checkmark  &         --  &         --  & \checkmark  & \checkmark \\
roll-up   & orthodox  &         --  &         --  & \checkmark  &         --  & \checkss{*} &         --  & \checkmark  & \checkmark \\
roll-up   & semisplit &         --  &         --  & \checkmark  &         --  &         --  &         --  & \checkmark  & \checkmark \\\hline
\end{tabular}
\begin{flushleft}
\textsuperscript{S} --\ equivalent to \emph{split} variant \\
\textsuperscript{J} --\ equivalent to \emph{join} variant \\
\textsuperscript{\bcancel{\checkmark}} --\ equivalent to another variant but less efficient, so this variant is useless\\
\textsuperscript{*} --\ works without using MAST in the expansion phase\\
\end{flushleft}
\end{center}\end{table}

Table~\ref{tab:heur_combinations} summarizes possible variants of action-based heuristics depending on the MCTS semisplit combination.
As a general rule, a variant is available if it gathers statistics required for the main MCTS phase where it is used (i.e., simulation phase for MAST and tree phase for RAVE).

The standard variants represent a straightforward application of the heuristics, i.e. when they operate exactly on the same (sub)moves as computed in MCTS.
They are available only when the domain of moves is the same in both the MCTS tree phases and the simulation phase, and they are always equivalent either to join or split variants.
Hence, we use the latter naming.

\mastsplit can work in all cases, as it can divide a move or its factor into a set of semimoves and supports evaluation of such a factor basing on the statistics of semimoves.
\ravesplit works only when the MCTS tree uses semisplit, as in other cases it could not easily evaluate submoves that are not single semimoves (due to technical difficulties and only small benefits expected, we do not propose such a solution).
\mastjoin and \ravejoin can work whenever the simulation phase and the tree phase, respectively, use the orthodox design.
\mastjoin can work when the MCTS tree phase is not orthodox, but then it is not applied for the expansion phase.
Finally, \mastcontext and \ravecontext can work in all cases, as they gather all statistics that may be required.
However, in some orthodox cases, they are fully equivalent to the corresponding join variants and thus are inefficient due to gathering some useless statistics (i.e., for semimoves).
Of course, the mix variants are available precisely in the combinations where both the split and the context variants are available, and where the context variant is non-optimal, they can also use the join variant instead (i.e., in the case of orthodox MCTS).

\clearpage
\section{Implementation in Regular Boardgames}\label{sec:rbg}

To test the concept in a practical setting and explain more issues in the context of a particular representation, we applied it to the Regular Boardgames (RBG) game description language.
It offers a simple way of partitioning a move, and furthermore, we can get significant efficiency benefits from it, resembling those that could be gained from a manual implementation of the reasoner.
Using a GGP language also allows us to split games in a more systematic way, without the need to take manual split decisions for each game.

\subsection{Regular Boardgames}

We give a short description of the language, which is necessary for understanding how exactly split strategies divide moves.
For the full language definition, we refer  to~\cite{Kowalski2019RegularBoardgames}.

To describe a game in  RBG we need to define a \emph{board}, \emph{variables}, \emph{player roles}, and \emph{rules}.
The \emph{game state} contains a placement of pieces on the board, values of the variables, information who is the current player, the current position (the one we are ``looking at'') on the board, and the current index (position) in the rules.
The \emph{board} is a directed graph with labeled edges, called \emph{directions}. (Thus, as any such graph is accepted, it is much more general than what is usually understood as a ``board''.)
The current player, in his turn, performs a sequence of elementary \emph{actions}. When applied sequentially, they modify the game state in a specific way. An action is \emph{legal} when it is: \emph{valid} for the current game state when it is applied; and permitted by the rules.
There are seven distinct types of actions:
\begin{enumerate}
\item \emph{Shift}, e.g., \lstinline|left| or \lstinline|up|. Changes the current position (the one we are looking at) on the board following the specified direction. The action is invalid when the edge specified is not defined in the board graph.
\item \emph{On}, e.g., \lstinline|{whiteQueen}|. Checks if the specified piece occupies the current position on the board. Does not modify the game state.
\item \emph{Off}, e.g., \lstinline|[whiteQueen]|. Puts the specified piece at the current position on the board.
The action is always valid and overrides the previous content of the square.
\item \emph{Comparison}, e.g., \lstinline|{$ turn==100}|. Compares two arithmetic expressions involving variables.
\item \emph{Assignment}, e.g., \lstinline|[$ turn=turn+1]|. Assigns the value of an arithmetic expression to a given variable.
\item \emph{Switch}, e.g., \lstinline|->white|. Ends a move and changes the current player to the one specified.
\item \emph{Pattern}, e.g., \lstinline|{? left up}|. A conditional valid only if there exists a legal sequence of actions under the specified rules. In this example, the action is valid if from the current square there is a path going on two edges labeled by \lstinline|left| and \lstinline|up|.
\end{enumerate}

A \emph{move} is defined as a sequence of actions ending with a switch.
\begin{example}\label{ex:amazons}
This sequence of actions originates from Amazons. It defines a move where a white queen is moving two squares up and then it shoots an arrow one square right.
{\rm\begin{lstlisting}[numbers=none]
{wQueen} [empty] up up [wQueen] right [arrow] ->black
\end{lstlisting}}
\end{example}
A \emph{move} is the subsequence of (indexed) actions (offs, assignments, and switches), paired with the positions in rules (i.e., regular expression) where they are applied.
These are precisely the actions that modify the game state, not counting the changes in the board position and the rules index.
Thus, the move given in the example above has length $4$ (caused by three offs and one switch).

A playout ends when the current player has no legal moves. Then, each player has assigned a \emph{score}, which is a value stored in a dedicated variable (named the same as the player's role).
The rules are encoded as a regular expression over the alphabet of the actions defined above.
The language defined by this expression encodes a game tree by containing all potentially legal sequences of actions.
To simplify the encoding of the regular expression, RBG descriptions may contain C-like macros that are instantiated for given parameters.

\subsection{Split Strategies}

The split strategies directly refer to the elements of the language.
The Mod component ends a semimove at every Off and Switch.
The Plus component additionally ends a semimove after entering an alternative of regular expressions. This, however, does not break consistent subexpressions with shifts only, which define the selection of a square.
The Shift component additionally provides a split point within such subexpressions after entering an alternative (similarly as Plus).

Split points in the generated game code rules are indicated with a dot (\lstinline|.|), whereas an Off and a Switch is always assumed to be a split point when semisplit is used.
Patterns are never split.

\subsubsection{Legality Check}
After the initial placement, we check whether the introduced split points do not potentially cause duplication of legal moves, i.e., whether we obtain a proper semisplit game.
It could happen that two distinct semimoves lead to the same legal move in the original rules.
For example, in \lstinline|(left* . + right* .) [e]|, if we stay at the same square, we could choose either the left branch or the right branch and end at different dots, but in the original rules the square of $[e]$ is the same, so it will lead to the same legal move.

Such illegal dots are postponed to the right until they become safe.
Thus, the above example will be turned into \lstinline|(left* + right* ).[e]|.

\subsubsection{Removing Indecisive Splits}
The third phase is optimization. We try to remove as many indecisive semimoves as possible.
An indecisive semimove is when it is the only outgoing semimove from a state.
In the above example, if we use the Mod component, the dot can be removed, as $[e]$ also ends a semimove.

Sometimes, we could remove either previous or the latter split point, but not both.
In this case, we remove split points starting from the right, thus making the first semimove shorter.
When it is impossible, because we cannot remove split points related to Offs and Switches when using Mod, we then remove other split points on the left.

Of course, not all indecisive semimoves are removed in this way, as whether a semimove is indecisive depends on the whole rules and is a computationally hard problem in general.

\subsection{Example of Amazons}

A complete example of game Amazons is given in Fig.~\ref{fig:amazons}.
Its underlying nondeterministic finite automaton, processed by the game compiler, is shown in Fig.\ref{fig:amazonsnfa}.
Split points introduced by each of the Mod, Plus, and Shift components are shown. A semimove ends when the current index in the rules reaches such a point.

At the beginning of the game description (Fig.~\ref{fig:amazons}, lines 1--14), we define the players, pieces, variables, and the game board.
For the player, we also provide their maximal achievable scores. In our example we need four pieces: \lstinline|e| for encoding an empty square, \lstinline|w| for a white queen, \lstinline|b| for a black one, and \lstinline|x| for an arrow. There are no variables here, as the variables for players containing their current scores are created automatically.
To define the board graph with its initial state, we use a predefined helper macro. We create a rectangular board with four possible movement directions.

Then, we introduce some helpful, user-defined macros to simplify further game definition. 
\lstinline|anySquare| is responsible for changing the current position to any square on board by first jumping an arbitrary number of squares vertically and then horizontally.
Split strategies with the Shift component insert a split point after the first such decision, whereas the second is skipped as it ends the shift subexpression.
The goal of \lstinline|directedShift| is to encode movement in direction \lstinline|dir| given as a parameter as long as the squares on the way are empty. It requires that at least one step has to be made.
\lstinline|queenShift| encodes all possible queen-like moves as an alternative to directed shifts. For diagonal movements, two consecutive directions are passed as a single macro argument. 

The \lstinline|turn| macro (lines 23--28), describes a single turn for player \lstinline|me|, whose queen pieces are encoded as \lstinline|piece|.
First, we switch the player to ourselves. Then, we switch the current square to any that contains our queen. 
Next, we pick up the queen -- putting down the ``empty'' piece, which introduces split under the Mod component.
Now, we select a direction, which forms a separate semimove under the Plus component.
Then, we move to the desired square and put down the queen (the second Mod split point).
This procedure is repeated for an arrow, as its placement follows the same rules as for moving a queen.
The switch action \lstinline|->>| gives control to the manager (special role named \emph{keeper}).
Keeper moves are never split; if there are split points inserted, they are ignored.
A keeper switch indicates that the player has no more decisions to make, and the remaining steps are deterministic in this sense.
These steps are: putting down the arrow and setting the winning score for the last player. When the playout ends because the current player has no legal moves, this stage will not be reached; thus the previous current player will win the game.
Finally, we can define the macro for the game rules. We encode the overall rules as a repetition of the sequence of the white and the black player turns (line~29).

\begin{figure}[ht]
\lstset{numbers=left,columns=fixed,xleftmargin=16pt}
\begin{lstlisting}
#players = white(100), black(100)
#pieces = e, w, b, x
#variables = // no variables
#board = rectangle(up,down,left,right,
         [e, e, e, b, e, e, b, e, e, e]
         [e, e, e, e, e, e, e, e, e, e]
         [e, e, e, e, e, e, e, e, e, e]
         [b, e, e, e, e, e, e, e, e, b]
         [e, e, e, e, e, e, e, e, e, e]
         [e, e, e, e, e, e, e, e, e, e]
         [w, e, e, e, e, e, e, e, e, w]
         [e, e, e, e, e, e, e, e, e, e]
         [e, e, e, e, e, e, e, e, e, e]
         [e, e, e, w, e, e, w, e, e, e])
#anySquare = ((up* + down*)(left* + right*))
#directedShift(dir) = (dir {e} (dir {e})*)
#queenShift = (
    directedShift(up left) + directedShift(up) +
    directedShift(up right) + directedShift(left) +
    directedShift(right) + directedShift(down left) +
    directedShift(down) + directedShift(down right)
  )
#turn(piece; me; opp) = (
    ->me anySquare {piece} [e]
    queenShift [piece]
    queenShift
    ->> [x] [$ me=100, opp=0]
  )
#rules = (turn(w; white; black) turn(b; black; white))*
\end{lstlisting}
\caption{RBG encoding of Amazons (non-split).}\label{fig:amazons}
\end{figure}

\clearpage
\begin{figure}[H]\begin{center}
\newcommand{\splitPointM}{\footnotesize\textbf{Mod}}
\newcommand{\splitPointP}{\footnotesize\textbf{Plus}}
\newcommand{\splitPointS}{\footnotesize\textbf{Shift}}
\begin{tikzpicture}[scale=1.3]\small
\tikzset{edge/.style = {->,> = latex}}
\tikzset{box/.style = {shape=rectangle,minimum size=1em,draw,dashed,fit=#1}}
\tikzset{sphere/.style = {shape=circle,minimum size=1em}}
\tikzset{vertex/.style = {shape=ellipse,minimum size=1em,draw}}

\node (main00) at (-1,-1) {};
\node[vertex] (main01) at (0,-1) {};
\node[vertex] (main02) at (0,-2) {};
\node[vertex] (main03) at (0,-4) {\splitPointS};
\node[vertex] (main04) at (0,-6) {};
\node[vertex] (main05) at (0,-6.65) {};
\node[vertex] (main07) at (0,-7.5) {\splitPointM};
\node[sphere] (QSdots) at (3,-10) {...};
\node[vertex] (main08) at (0,-13) {};
\node[vertex] (main085) at (-0.45,-13.8) {\splitPointM};
\node[vertex] (main09) at (-3,-14.5) {};
\node[vertex] (main10) at (-2.25,-14.5) {};
\node[vertex] (main11) at (0.25,-14.5) {};
\node[vertex] (main12) at (2.25,-14.5) {};
\node (main13) at (4.1,-14.5) {};
\node (main14) at (4.1,-2) {};

\node[vertex] (up1) at (1,-2.5) {};
\node[vertex] (up2) at (1,-3.5) {};

\node[vertex] (down1) at (-1,-2.5) {};
\node[vertex] (down2) at (-1,-3.5) {};

\node[vertex] (left1) at (1,-4.5) {};
\node[vertex] (left2) at (1,-5.5) {};

\node[vertex] (right1) at (-1,-4.5) {};
\node[vertex] (right2) at (-1,-5.5) {};

\draw[dashed] (-3.3,-7.25) rectangle (3.3,-13.25);
\node (QStitle) at (-2.4,-7.5) {$\mathrm{queenShift}$};

\coordinate (QSboxA) at (-2.4,-13.55);
\coordinate (QSboxB) at (0.0,-14.05);
\node[box={(QSboxA) (QSboxB)}] (QSbox) {};
\node at (-1.6,-13.8) {$\mathrm{queenShift}$};

\node[vertex] (QSl1) at (-1,-8.5) {\splitPointP};
\node[vertex] (QSl2) at (-1,-9.5) {};
\node[vertex] (QSl3) at (-1,-10.5) {};
\node[vertex] (QSl4) at (-1,-11.5) {};
\node[vertex] (QSl5) at (-2,-11.5) {};
\node[vertex] (QSl6) at (-3,-11.5) {};
\node[vertex] (QSl7) at (-3,-12.5) {};
\node[vertex] (QSl8) at (-2,-12.5) {};

\node[vertex] (QSu1) at (0,-8.5) {\splitPointP};
\node[vertex] (QSu3) at (0,-9.5) {};
\node[vertex] (QSu4) at (0,-10.5) {};
\node[vertex] (QSu5) at (1,-10.5) {};
\node[vertex] (QSu7) at (1.75,-11) {};
\node[vertex] (QSu8) at (1,-11.5) {};

\draw[dotted] (-3.55,-1.6) rectangle (3.55,-15.1);
\node[rotate=90,anchor=north] (white) at (-3.55,-3.15) {$\mathrm{turn(w; white; black)}$};

\draw[dotted] (4.0,-1.6) rectangle (5.0,-15.1);
\node[rotate=90,anchor=north] (white) at (4.0,-3.15) {$\mathrm{turn(b; black; white)}$};

\draw[edge] (main00) edge (main01);
\draw[edge] (main01) edge node[above right]{$\mathit{{\to}white}$} (main02);
\draw[edge] (main02) edge node[right]{$\mathit{\varepsilon}$} (main03);
\draw[edge] (main03) edge node[right]{$\mathit{\varepsilon}$} (main04);
\draw[edge] (main04) edge node[right]{$\mathit{\{w\}}$} (main05);
\draw[edge] (main05) edge node[right]{$\mathit{[e]}$} (main07);
\draw[edge] (main07) edge[bend left = 19] (QSdots);
\draw[edge] (main07) edge[bend left = 27] (QSdots);
\draw[edge] (main07) edge[bend left = 35] (QSdots);
\draw[edge] (main07) edge[bend left = 44] (QSdots);
\draw[edge] (main07) edge[bend left = 54] (QSdots);
\draw[edge] (main07) edge[bend left = 65] (QSdots);
\draw[edge] (QSdots) edge[bend left = 19] (main08);
\draw[edge] (QSdots) edge[bend left = 27] (main08);
\draw[edge] (QSdots) edge[bend left = 35] (main08);
\draw[edge] (QSdots) edge[bend left = 44] (main08);
\draw[edge] (QSdots) edge[bend left = 54] (main08);
\draw[edge] (QSdots) edge[bend left = 65] (main08);
\draw[edge] (main08) edge[bend left] node[right]{$\mathit{[w]}$} (main085);
\draw[edge] (QSbox) edge[bend right] node[left]{$\mathit{\to\mathrel{\mkern-13mu}\to}$} (main09);
\draw[edge] (main09) edge node[below]{$\mathit{[x]}$} (main10);
\draw[edge] (main10) edge node[below]{$\mathit{[\$white{=}100]}$} (main11);
\draw[edge] (main11) edge node[below]{$\mathit{[\$black{=}0]}$} (main12);
\draw[edge] (main12) edge node[below]{$\mathit{{\to}black}$} (main13);
\draw[edge] (main14) edge node[below]{$\mathit{{\to}white}$} (main02);

\draw[edge] (main02) edge[bend right] node[above]{$\mathit{\varepsilon}$} (up1);
\draw[edge] (up1) edge[bend left] node[right]{$\mathit{up}$} (up2);
\draw[edge] (up2) edge[bend left] node[left]{$\mathit{\varepsilon}$} (up1);
\draw[edge] (up2) edge[bend right] node[below right]{$\mathit{\varepsilon}$} (main03);

\draw[edge] (main02) edge[bend left] node[above]{$\mathit{\varepsilon}$} (down1);
\draw[edge] (down1) edge[bend right] node[left]{$\mathit{down}$} (down2);
\draw[edge] (down2) edge[bend right] node[right]{$\mathit{\varepsilon}$} (down1);
\draw[edge] (down2) edge[bend left] node[below left]{$\mathit{\varepsilon}$} (main03);

\draw[edge] (main03) edge[bend right] node[above]{$\mathit{\varepsilon}$} (left1);
\draw[edge] (left1) edge[bend left] node[right]{$\mathit{left}$} (left2);
\draw[edge] (left2) edge[bend left] node[left]{$\mathit{\varepsilon}$} (left1);
\draw[edge] (left2) edge[bend right] node[below right]{$\mathit{\varepsilon}$} (main04);

\draw[edge] (main03) edge[bend left] node[above]{$\mathit{\varepsilon}$} (right1);
\draw[edge] (right1) edge[bend right] node[left]{$\mathit{right}$} (right2);
\draw[edge] (right2) edge[bend right] node[right]{$\mathit{\varepsilon}$} (right1);
\draw[edge] (right2) edge[bend left] node[below left]{$\mathit{\varepsilon}$} (main04);

\draw[edge] (main07) edge[bend right] node[above left]{$\mathit{\varepsilon}$} (QSl1);
\draw[edge] (QSl1) edge node[left]{$\mathit{up}$} (QSl2);
\draw[edge] (QSl2) edge node[left]{$\mathit{left}$} (QSl3);
\draw[edge] (QSl3) edge node[left]{$\mathit{\{e\}}$} (QSl4);
\draw[edge] (QSl4) edge node[left]{$\mathit{\varepsilon}$} (main08);
\draw[edge] (QSl4) edge node[below]{$\mathit{\varepsilon}$} (QSl5);
\draw[edge] (QSl5) edge[bend right] node[above]{$\mathit{up}$} (QSl6);
\draw[edge] (QSl6) edge[bend right] node[right]{$\mathit{left}$} (QSl7);
\draw[edge] (QSl7) edge[bend right] node[below]{$\mathit{\{e\}}$} (QSl8);
\draw[edge] (QSl8) edge[bend right] node[left]{$\mathit{\varepsilon}$} (QSl5);
\draw[edge] (QSl8) edge node[above]{$\mathit{\varepsilon}$} (main08);

\draw[edge] (main07) edge node[right]{$\mathit{\varepsilon}$} (QSu1);
\draw[edge] (QSu1) edge node[right]{$\mathit{up}$} (QSu3);
\draw[edge] (QSu3) edge node[right]{$\mathit{\{e\}}$} (QSu4);
\draw[edge] (QSu4) edge node[right]{$\mathit{\varepsilon}$} (main08);
\draw[edge] (QSu4) edge node[below]{$\mathit{\varepsilon}$} (QSu5);
\draw[edge] (QSu5) edge[bend left = 30] node[above]{$\mathit{up}$} (QSu7);
\draw[edge] (QSu7) edge[bend left = 30] node[below]{$\mathit{\{e\}}$} (QSu8);
\draw[edge] (QSu8) edge[bend left] node[right]{$\mathit{\varepsilon}$} (QSu5);
\draw[edge] (QSu8) edge node[above]{$\mathit{\varepsilon}$} (main08);
\end{tikzpicture}
\caption{NFA represented by the Amazons description from Fig.~\ref{fig:amazons}.}\label{fig:amazonsnfa}
\end{center}\end{figure}
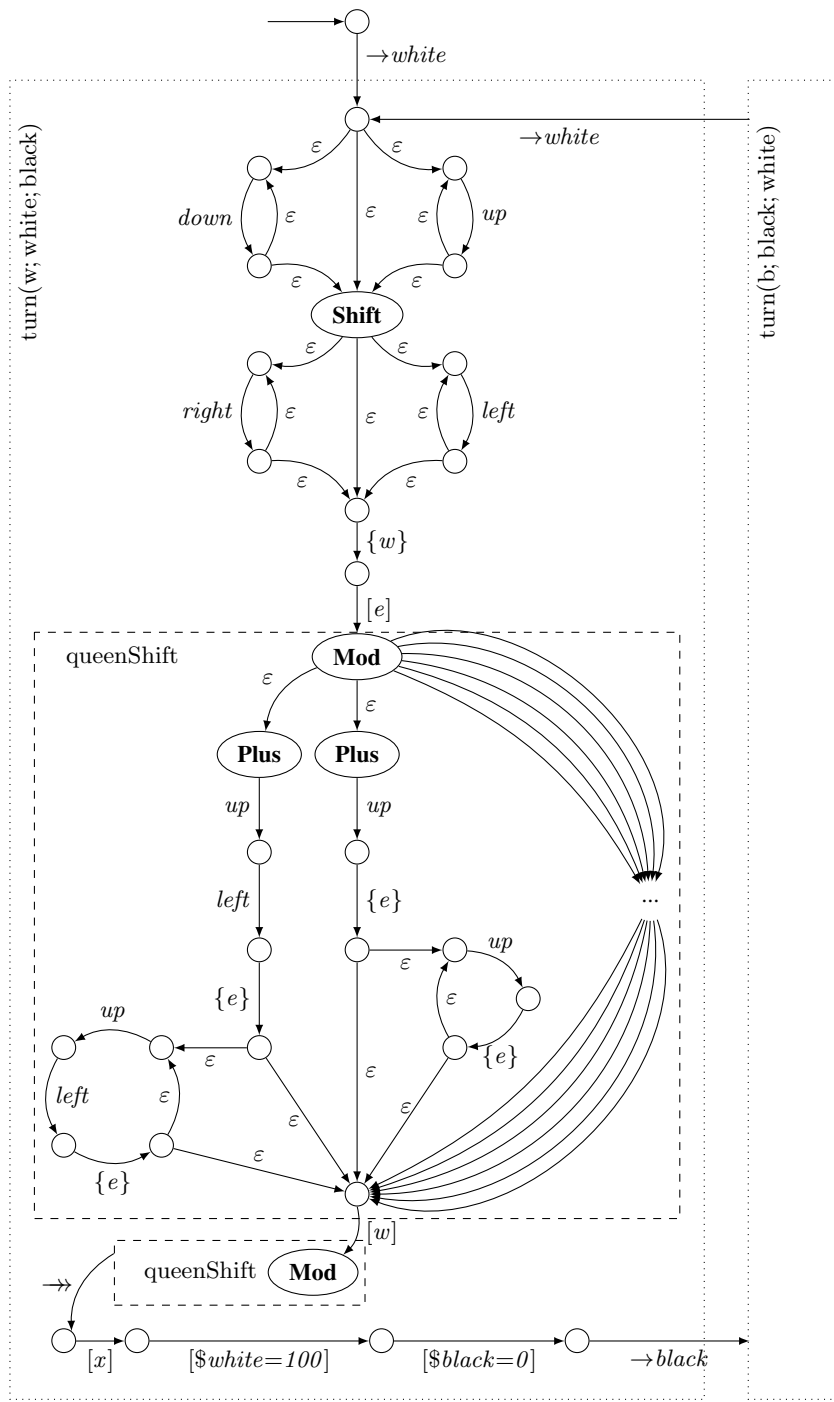

\clearpage
\section{Description of Games in the Test Set}\label{sec:games}

We describe the rules of all games used in our experiments and how they are split according to the three considered components of our split strategies.
All these games are two-player, perfect-information, and deterministic.

For all games, the Shift component separates choosing a column (first) and a row (second) when selecting any square, which usually takes place at the beginning of a turn.
All the strategies sometimes introduce some indecisive semimoves, which we do not describe in detail.
They can slightly affect efficiency, but generally are irrelevant for the quality of MCTS iterations.
They come from the fact that the splitting algorithm cannot know whether a split point is decisive, as it only splits moves basing on syntactic game rules.

For each game, it is given the name of the game file in the provided Code Appendix.

\noindent\emph{$\bullet$ Amazons} (\texttt{amazons.rbg}):
It is played on a $10\times 10$ chessboard. The initial position contains four pieces called \emph{amazons} for each player. The move consists of shifting an own amazon, which moves as a Chess queen, to a valid empty square, and then, from its final position, spawning a new piece called an arrow that has to be shifted under the same conditions.  Arrows block the squares where they are put for the remaining part of the play. Neither an amazon nor an arrow can be shifted through a square with an arrow or another amazon.
No captures are possible, and the player without a legal move loses. Draws are impossible.

Amazons is considered to be multi-action because moving a queen and shooting an arrow can be performed in separate moves.
In particular, a moved queen always has a possibility to shot an arrow; thus this would yield a proper split-equivalent game.

The Mod component divides a move into three choices ending by modifying a square: taking an amazon, putting it on a destination square, and putting an arrow.
The Plus component additionally separates the choice of the eight directions for both an amazon and an arrow.

\noindent\emph{$\bullet$ Breakthrough} (\texttt{breakthrough.rbg}):
The game is played on the standard chessboard, starting with two rows of pawns on each side of the board. Pawns can capture diagonally and go forward either vertically or diagonally. The goal of the game is to reach the opposite row. Draws are impossible.

The Mod component simply separates piece selection and its destination square (maximum three possibilities).
Here, the Plus component does not introduce any additional decisive semimoves.

\noindent\emph{$\bullet$ Breakthru} (\texttt{breakthru.rbg}): 
This is a game of evasion or capture, played on $11\times 11$ board featuring unevenly matched teams with different objectives.
In the initial phase, the \emph{gold} player puts his pieces in the central part of the board with the goal of escorting a distinguished piece called flagship to the perimeter of the board.
The other \emph{silver} player puts his ships on predefined positions in the outward part of the board and tries to capture the flagship.
In one turn, a player may perform two moves of a standard unit or one move with a flagship (both behave like Chess rook without capture), or one capture. When capturing, a piece behaves like a Chess bishop limited to one square.

The Mod component works as usual, yet the player already decides together with grabbing the first piece whether to capture or not.
With the Plus component, this decision of capturing takes place separately after selecting the square, since there is an alternative on these choices.
Furthermore, choosing one of the four directions is also performed separately.
With Shift, besides separate selection of a column and a row, a diagonal shift is also split into up-down and left-right decisions.
The initial phase of the gold player's setup is split only under the Shift component.

\noindent\emph{$\bullet$ Chess} (\texttt{chess.rbg}):
The Chess version with the fifty-move rule for a draw. It contains all the standard rules (check, castling, en-passant, etc.), except the threefold position repetition checking.

Chess follows the standard pattern for splitting.
Thus, the Mod component separates choosing the piece and its destination square.
There are two exceptions for rare moves: castling, which is decided after picking the king; and en-passant, which is decided at the beginning of a move if available.
An additional notable efficiency improvement comes from postponed verification of whether the move is legal, i.e., whether after the move, the king is not under check.
With Plus, picking a direction for Bishop, Rook, and Queen is also separated, together with castling direction.
Shift also divides King and Knight moves (e.g., a knight first decides about the 2-squares hop direction, then the 1-square hop).

\noindent\emph{$\bullet$ Chess without check} 
(\texttt{chess\_kingCapture.rbg}):
This is a simplified version of Chess, commonly used in GGP (its GDL counterpart is called \texttt{speedchess}).
There is no check, i.e., it is ignored in all cases, and the game ends with capturing the king.
Strategically, it is the same as Chess, except that there is no stalemate.

Splitting is the same as for Chess, but of course, we do not have any postponed check checking.

\noindent\emph{$\bullet$ English draughts} (\texttt{englishDraughts.rbg}):
It is one of the most popular versions of draughts played on an $8\times 8$ board.
A crowned man (promoted after reaching the opposite row) can move one square both forward or backward.
Performing capturing moves are mandatory whenever possible, but the player is not forced to perform a longest sequence of captures as in other draughts versions.
The game ends in a draw after 20 moves without capturing or moving a (non-crowned) man.

The Mod component divides a capture into picking a piece, removing the opponent (indecisive semimove), and putting a piece.
Similarly, a non-capturing move is separated into two semimoves.
The Plus component additionally separates picking a direction, but, as this already determines the destination square, this is only an indecisive semimove.
Shift works as usual, yet due to a different shape of the board (black squares only), columns and rows are perceived as diagonals.

\noindent\emph{$\bullet$ Fox and Hounds} (\texttt{foxAndHounds.rbg}):
Also known as \emph{Wolf and Sheep}, the game is played on an $8\times 8$ draughts-like board (i.e., only black squares are used).
Initially, four hounds occupy one row of the board, and the fox player places his piece on any black square of the opposite row. The hounds move like a draughts man (diagonally forward one square) while the fox moves like a draughts king (diagonally forward or backward one square).
The fox moves first and chooses its initial square.
The players move one of their pieces alternatingly.
The winning condition for the fox is to reach the hounds' back rank, and the player that is unable to move loses.

Splitting here follows the standard pattern.
Since there is just one fox, Mod affects only hound moves.
Plus does not introduce any more decisive moves.

\noindent\emph{$\bullet$ Go} (\texttt{go\_constsum.rbg}):
This is a $19\times 19$ Go implementation using Tromp-Taylor rules with area scoring, suicide prohibition, no ko rule, and turn limit set to 722.
This is a constant-sum version with win/draw/loss results based on area scoring.

Here, a move is split into the decision whether to pass (set pass variable) and then, if not passing, the selection of a square.
This game is a distinguished case demonstrating how a particular encoding significantly decides about splits.
The Mod component separates the decision about the passing and selecting a square.
Plus does not add any more decisive semimoves, and Shift separates column and row selection as usual.

\noindent\emph{$\bullet$ Knightthrough} (\texttt{knightthrough.rbg}): 
The game, originated from Stanford's GGP competition, is very similar to Breakthrough. The pawns are replaced by Chess knights that can only move forward.

Splitting is the same as for Breakthrough.

\noindent\emph{$\bullet$ Pentago} (\texttt{pentago.rbg}):
This is a five-in-a-row game played on a $6\times 6$ board with the additional rule that after every move the current player has to rotate by 90 degrees on of the four $3\times 3$ subboards. The winning condition is checked twice per turn: before and after the rotation.

The Mod component separates putting a ball and rotating a subboard.
Plus additionally separates the decision of the rotation direction (thus in ModPlus, there are three semimoves in a turn).
Shift just separates column and row selection, which affects both putting a ball and selecting one of the subboards (i.e., first up-down, then left-right subboard).

\noindent\emph{$\bullet$ Skirmish} (\texttt{skirmish.rbg}):
Yet another simplified version of Chess, also introduced by Stanford's GGP competition.
The initial position and movement rules are the same as in Chess (including promotion, castling, and en-passant), but the king is treated as a regular piece, which can be captured as the others.
The game ends when a player does not have a legal move, or 100 turn limit is reached.
The player with more pieces left wins (each piece counts as 1).

Splitting is the same as for Chess without check.

\noindent\emph{$\bullet$ The Mill Game} (\texttt{theMillGame.rbg}):
Also known as \emph{Nine men's morris}. 
The game is played on a grid with twenty-four intersections. Each player has nine pieces called men. Their goal is to form mills: three own men lined horizontally or vertically, which allows removing the opponent's men.

The game has three phases: placing men on vacant points, 
moving men to adjacent points, and optionally, moving men to any vacant point when the player already has three men only.
A player loses when it is left with less than three pieces or has no legal moves.
The game ends with a draw after 20 moves without forming a mill after the first phase.

The Mod component separates a pawn movement as usual, and also capturing; thus a turn consists of either two or three semimoves.
Plus separates choosing a direction in a standard way.
Due to a non-standard board, the square selection with Shift also works differently: first one of the three rings is chosen, then a square on this ring.
This affects both own pawn selection and capturing.

\clearpage
\section{Experimental Setup}

\subsection{Parameters}\label{subsec:parameters}

All parameters of the tested agents were turned according to the recommendations in the literature.
Both orthodox and semisplit agents were using exactly the same parameters set.
This means that they may not necessarily be tuned for semisplit agents, as they are originally recommended for orthodox agents in the literature.

The exploration constant in the UCT formula \cite[(1)]{sironi2016comparison} was set to $C=0.4$ \cite{Finnsson2010Learning}.
We have also tried the second popular value $C=0.7$, but it turned out to give worse results than $C=0.4$ for our setting and game set.

MAST uses the $\varepsilon$-greedy policy with $\varepsilon=0.4$, i.e., with probability $0.6$ it chooses a (sub)move with the best average score.
The decaying parameter was set to $0.2$ \cite{tak2014decaying}.

In the case of \mastmix, the mix-threshold was set to $7$, as this value was proposed for NST \cite{tak2012ngrams}.

RAVE had its equivalence parameter set to $250$ \cite{sironi2016comparison}.
According to recommendations, whenever RAVE was used (in all agent types), the exploration constant was also set to $C=0.2$.

An agent does not compute (it sleeps) during the opponent's turn.

All ties (e.g., moves with the same UCT score or MAST score) are broken uniformly at random.
In the final selection, an agent selects the move/semimove with the largest average score, and in the case of ties, with the largest number of iterations (and then ties are resolved randomly).

\subsection{Algorithmic Details}

MAST stores global statistics for each move.
Updating and, in particular, reading those statistics should be as fast as possible to avoid the situation in which the time overhead from using MAST causes losings greater than benefits.

Probably the best general choice is a dictionary based on a hashtable.
We use a custom implementation of a hashtable with closed hashing, taking into account its global character (adjusting the initial size and growth) and the domain of the integers in actions.
However, in particular in the case of \mastsplit, when we know that the domain of all possible semimoves is small, the hashtable can be replaced with a raw array.

\mastcontext could be implemented with the same hashtable as is used for full moves.
However, it can be optimized, taking advantage of our specific application.
We use the fact that whenever a submove is stored in the hashtable, also all its prefixes must have been already stored there since they have been applied earlier.
Furthermore, we have queried these prefixes right before processing the submove.
Thus, instead of the full submove, we store pairs of the context and the semimove.
The context is represented by the bucket id in the hashtable, which uniquely represents a stored submove and is returned directly from the preceding query.
In this way, elements in the hashtable have a fixed length, and their hashing and comparisons (in the case of conflicts) are cheaper.

RAVE uses similar structures, but it is enough to use a hash set instead of a hash map, as we only need to query if a given move occurs at the bottom of the iteration.
In the case of \ravesplit, we use bit sets to encode this, taking advantage of a relatively small domain of semimoves.

When MAST is used in a semisplit simulation, we compute all values of semimoves only once, using a heap (priority queue) for picking up the best value.
This reduces the worst-case time from quadratic to linear-logarithmic in the number of available semimoves, when all semimoves turn out to be dead.

\subsection{The Setting}

There are 300 plays in each test, where 150 plays are played with swapped agents.
The final win ratio is computed by assigning $1$ point for a win, $0.5$ for a draw, and $0$ for a loss.
Win ratios are given in percents.

The confidence intervals are given for $95\%$ confidence based using a standard method \cite{haworth2003self}.
They are used to count games where an agent has a clear advantage or a disadvantage, as described in the next section.
(The confidence interval is not given in the case of winning all the plays by one side, which sometimes happens for Breakthru.)

In the timed experiments (\emph{timed setting}), the time limit is set to $0.5$s per turn, unless otherwise stated.
The buffer time is not included, i.e., this is the time limit for pure computation; the real limit sent by the manager is $0.6s$.
In view of computational power, this roughly corresponds to the setting with $10$s limit based on GDL reasoning \cite{Kowalski2020EfficientReasoning}, which is often used for comparisons of agents \cite{sironi2016comparison}.

There are also experiments with a fixed budget limit (\emph{fixed setting}).
We limit the number of states that an agent can compute in its own turn.
Only nodal states are counted.
The limits for each game are computed by benchmarking the corresponding orthodox agent, measuring the number of states computed in the first turn over $10$s.
In contrast with limiting the number of simulations, this better corresponds to the timed setting, because of two reasons.
First, the average length of simulations may vary depending on whether the semisplit or orthodox design is used.
Second, the average length of simulations decreases as gameplay advances, since terminal states are closer.

\subsection{Environment}

The hardware used for the timed experiments was a computer with AMD Ryzen Threadripper 3970X 32-Core Processor and 64GBi.
The system was \texttt{Ubuntu 20.04.2 LTS (GNU/Linux 5.4.0-80-generic x86\_64)}.
The relevant software was \texttt{gcc 9.3.0}, \texttt{boost 1.75.0}, and \texttt{Python 3.8.10}.

The tests of agents were performed with 30 plays in parallel (one process computes 5 plays, giving 150 in total for one side).

Because the experiments with fixed states limit do not depend on hardware efficiency (i.e., are deterministic up to the random seed), they were performed on other computers belonging to the same grid.

The random generator used was the default used in the RBG system: \texttt{std::mt19937} together with the Lemire's selection method \cite{Lemire2018FastRandom,Kowalski2020EfficientReasoning}.

\clearpage
\section{Extended Results}

We present extended results in addition to those given in the main text.

In all tables, the bars show the grand average in timed (upper, red) and fixed (lower, blue) settings. Timed results are also given separately for each game with 95\% confidence intervals.
Green, yellow, and red indicate statistically significant \emph{better}, \emph{same}, and \emph{worse} performance, respectively.
A game is counted for better performance if the whole 95\% confidence interval is above 50\%.
Worse performance is symmetrical (i.e., the confidence interval is below 50\%), and the draw means that 50\% is included within the confidence interval.

\subsection{Vanilla Agents}

Tables~\ref{tab:agent_split_strategies1}, \ref{tab:agent_split_strategies2}, and ~\ref{tab:agent_split_strategies3} show the results for different split strategies.
We have three types of agents combined with three split strategies of different granularity, all larger than that of Mod alone.
As general observation, we conclude that the Shift component, which commonly separates board's column and row selection, tends to decrease the results, but the Plus component has a small positive effect.

\begin{table}[!htb]\renewcommand{\arraystretch}{0.5}
\newcommand{\colTwo}[1]{\multicolumn{2}{c|}{#1}}
\newcommand{\colBTwo}[1]{\multicolumn{2}{|c|}{#1}}
\newcommand{\wid}[0]{230px}
\newcommand{\colG}[1]{\includegraphics[
width=\wid]{figures_winrates/#1.pdf}}
\newcommand{\notbot}[1]{\vbox{\hbox{\strut #1}\hbox{\strut \vphantom{a}} }}
\caption{Win rates of semisplit agents with different split strategies.
}\label{tab:agent_split_strategies1}
\begin{center}\begin{tabular}{|l|c|}\hline
\multicolumn{1}{|c|}{Agent}   &  Win rates vs. \orthodox\phantom{\Large X} \\[3pt]\hline
\notbot{\semisplitV{S-nodal}{S}{ModPlus}} & \colG{MP_Sn_S} \\ \hline
\notbot{\semisplitV{S-raw}{S}{ModPlus}} & \colG{MP_Sr_S} \\ \hline
\notbot{\semisplitV{S-nodal}{S}{ModShift}} & \colG{MS_Sn_S} \\ \hline
\notbot{\semisplitV{S-nodal}{S}{ModPlusShift}} & \colG{MPS_Sn_S} \\ \hline
&  \includegraphics[angle=90,width=\wid]{figures_winrates/gamenames.pdf}\\\hline
\end{tabular}\end{center}
\end{table}
\begin{table}[!htb]\renewcommand{\arraystretch}{0.5}
\newcommand{\colTwo}[1]{\multicolumn{2}{c|}{#1}}
\newcommand{\colBTwo}[1]{\multicolumn{2}{|c|}{#1}}
\newcommand{\wid}[0]{230px}
\newcommand{\colG}[1]{\includegraphics[
width=\wid]{figures_winrates/#1.pdf}}
\newcommand{\notbot}[1]{\vbox{\hbox{\strut #1}\hbox{\strut \vphantom{a}} }}
\caption{Win rates of semisplit orthodox-tree design agents with different split strategies.
}\label{tab:agent_split_strategies2}
\begin{center}\begin{tabular}{|l|c|}\hline
\multicolumn{1}{|c|}{Agent}   &  Win rates vs. \orthodox\phantom{\Large X} \\[3pt]\hline
\notbot{\semisplitV{O}{S}{ModPlus}} & \colG{MP_O_S} \\ \hline
\notbot{\semisplitV{O}{S}{ModShift}} & \colG{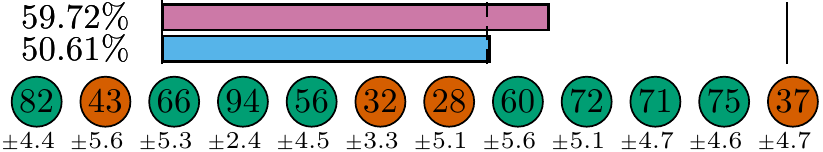} \\ \hline
\notbot{\semisplitV{O}{S}{ModPlusShift}} & \colG{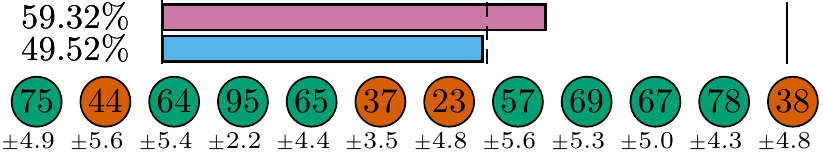} \\ \hline
&  \includegraphics[angle=90,width=\wid]{figures_winrates/gamenames.pdf}\\\hline
\end{tabular}\end{center}
\end{table}
\begin{table}[!htb]\renewcommand{\arraystretch}{0.5}
\newcommand{\colTwo}[1]{\multicolumn{2}{c|}{#1}}
\newcommand{\colBTwo}[1]{\multicolumn{2}{|c|}{#1}}
\newcommand{\wid}[0]{230px}
\newcommand{\colG}[1]{\includegraphics[
width=\wid]{figures_winrates/#1.pdf}}
\newcommand{\notbot}[1]{\vbox{\hbox{\strut #1}\hbox{\strut \vphantom{a}} }}
\caption{Win rates of semisplit roll-up agents with different split strategies.
}\label{tab:agent_split_strategies3}
\begin{center}\begin{tabular}{|l|c|}\hline
\multicolumn{1}{|c|}{Agent}   &  Win rates vs. \orthodox\phantom{\Large X} \\[3pt]\hline
\notbot{\semisplitV{R-nodal}{S}{ModPlus}}   & \colG{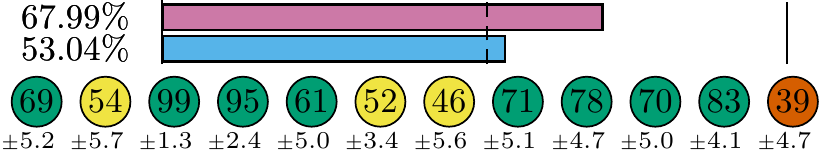} \\ \hline
\notbot{\semisplitV{R-nodal}{S}{ModShift}}   & \colG{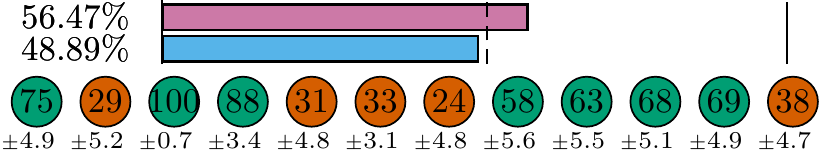} \\ \hline
\notbot{\semisplitV{R-nodal}{S}{ModPlusShift}}   & \colG{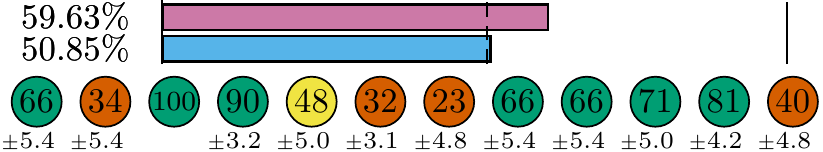} \\ \hline
&  \includegraphics[angle=90,width=\wid]{figures_winrates/gamenames.pdf}\\\hline
\end{tabular}\end{center}
\end{table}

\clearpage
\subsection{Enhanced Agents}

Table~\ref{tab:agents_orthodox_heur} shows the results from sanity tests concerning how beneficial is adding the action-based heuristics to the vanilla orthodox agent in our setting.
We observe that both heuristics give some advantage on average, yet MAST yields a larger improvement.
However, the results are diminished by the time overhead, which is not surprising as both heuristics involve noticeable computational cost (especially when computing moves and states is fast).

\begin{table}[!htb]\renewcommand{\arraystretch}{0.5}
\newcommand{\colTwo}[1]{\multicolumn{2}{c|}{#1}}
\newcommand{\colBTwo}[1]{\multicolumn{2}{|c|}{#1}}
\newcommand{\wid}[0]{230px}
\newcommand{\colG}[1]{\includegraphics[width=\wid]{figures_winrates/#1.pdf}}
\newcommand{\colGg}[1]{\includegraphics[width=\wid]{figures_winrates/#1.pdf}}
\newcommand{\notbot}[1]{\vbox{\hbox{\strut #1}\hbox{\strut } }}
\caption{Win rates of orthodox agents with action-based heuristics vs. vanilla orthodox agent.}\label{tab:agents_orthodox_heur}
\begin{center}\begin{tabular}{|l|c|}\hline
\multicolumn{1}{|c|}{Agent}   &  Win rates vs. \orthodox\phantom{\Large X} \\[3pt]\hline
\notbot{\orthodoxM} & \colG{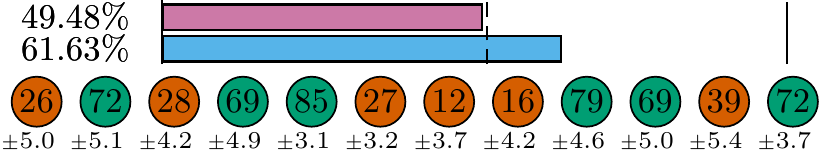} \\ \hline
\notbot{\orthodoxR} & \colG{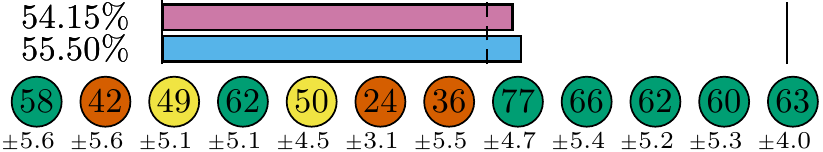} \\ \hline
\notbot{\orthodoxMR} & \colG{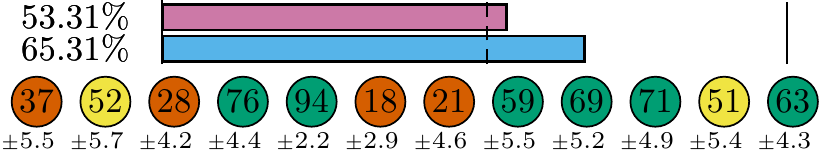} \\ \hline
&  \includegraphics[angle=90,width=\wid]{figures_winrates/gamenames.pdf}\\\hline
\end{tabular}\end{center}
\end{table}

\begin{figure*}[htb!]\centering
\includegraphics[width=\textwidth]{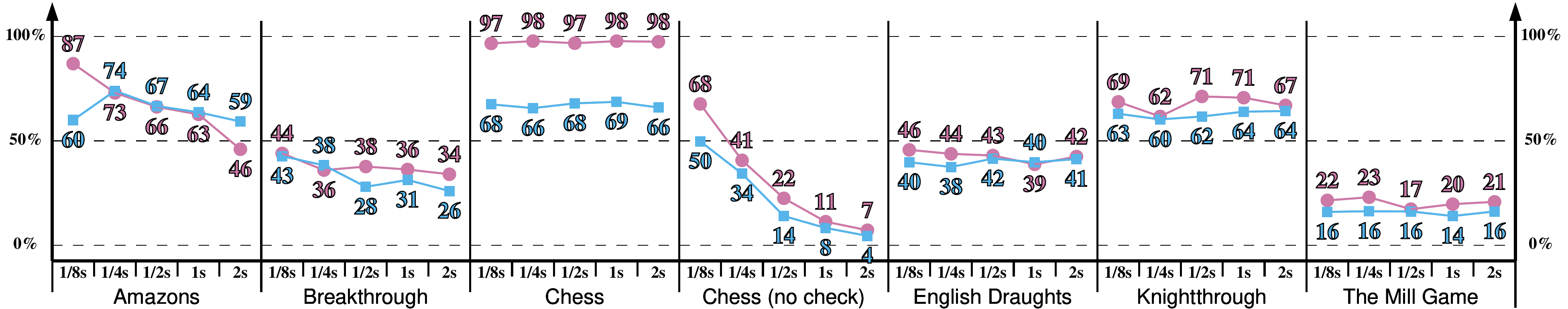}
\caption{The results of \semisplit{S-nodal,\ravesplit}{S,\mastsplit}{Mod} versus \orthodoxMR for different time limits (circle, red) and for equivalent states budgets (square, blue).}
\label{fig:experiments-heur}
\end{figure*}

Figure~\ref{fig:experiments-heur} is the counterpart to the Figure~1 from the main text but with enhanced MCTS agents.
Here, we show the results for the most straightforward (except that this is nodal) semisplit variant.
The tendencies are generally similar, and we get comparable results across different time limits, yet in a few cases (Amazons, Breakthrough, and Chess (no check)) they drop as the time limit grows.
The differences between the results in timed and fixed settings are also smaller, except for Chess, where computing legal moves is costly.
It is because action-based heuristics are expensive relatively to semisplit's MCTS computation.
This should motivate us to consider more involved variants when semisplit is combined with these heuristics.

Tables~\ref{tab:agents_heur1}, \ref{tab:agents_heur2}, and \ref{tab:agents_heur3} show different variants of semisplit agents together with some combinations of action-based heuristics. Due to the number of all possible combinations, we had to make some selection.
We tested agents that performed best/most robust in the vanilla case, i.e., \semisplit{S-nodal}{S}{Mod}, \semisplit{O}{S}{Mod}, and \semisplit{R-nodal}{S}{Mod}, which are shown in the three tables, respectively.
We also kept Mod split strategy as the simplest choice, since adding the Shift component seems to give worse results and adding Plus does not change much.

First, we observe that the split variants of the heuristics give the best results in the timed setting.
\mastmix gives a slight improvement in quality, but this benefit is discounted by time overhead.
It is clearly better than \mastcontext (compare \semisplit{S-nodal}{S,\mastcontext}{Mod} with \semisplit{S-nodal}{S,\mastmix}{Mod}, and \semisplit{O,\ravejoin}{S,\mastcontext}{Mod} with \semisplit{O,\ravejoin}{S,\mastmix}), whereas \mastsplit is faster enough to get better results in the timed setting.

When semisplit design is applied in the MCTS tree, RAVE gives worse results, also in the fixed setting.
A reason for that could be that semimoves carry too little information, so the hints gave by RAVE are too random.
We should take into account that RAVE gives only little benefits also for orthodox agents.
Using RAVE also means setting the exploration constant $C$ to $0.2$ instead of $0.4$ (Subsection~\ref{subsec:parameters}), and since semisplit also reduces exploration, this can be too much.
Further experiments concerning this particular case may be required, and this is a topic for future work.
However, RAVE still gives a slight advantage in the fixed setting when applied in MCTS tree under orthodox design (compare \semisplit{O}{S,\mastsplit}{Mod} and \semisplit{O,\ravejoin}{S,\mastsplit}{Mod}).

The best performing and most robust variants are \semisplit{O}{S,\mastsplit}{Mod} and \semisplit{R-nodal}{S,\mastsplit}{Mod}.
The latter with RAVE added gets slightly better results in the fixed setting.
Recall here that \ravecontext is the only available variant that works with the roll-up MCTS variant.

\begin{table}[!htb]\small\renewcommand{\arraystretch}{0.1}
\newcommand{\colTwo}[1]{\multicolumn{2}{c|}{#1}}
\newcommand{\colBTwo}[1]{\multicolumn{2}{|c|}{#1}}
\newcommand{\wid}[0]{210px}
\newcommand{\colG}[1]{\includegraphics[width=\wid]{figures_winrates/#1.pdf}}
\newcommand{\notbot}[1]{\vbox{\hbox{\strut #1}\hbox{\strut \vphantom{a}} }}
\caption{Win rates of semisplit agents based on \semisplit{S-nodal}{S}{Mod} with variants of action-based heuristics.}\label{tab:agents_heur1}
\begin{center}\begin{tabular}{|l|c|}\hline
\multicolumn{1}{|c|}{Agent}     &  Winrates vs. \orthodoxMR\phantom{\LARGE X} \\[4pt]\hline
\notbot{\semisplitV{S-nodal,\ravesplit}{S,\mastsplit}{Mod}} & \colG{M_SnRs_SMs} \\ \hline
\notbot{\semisplitV{S-nodal,\ravecontext}{S,\mastcontext}{Mod}}  & \colG{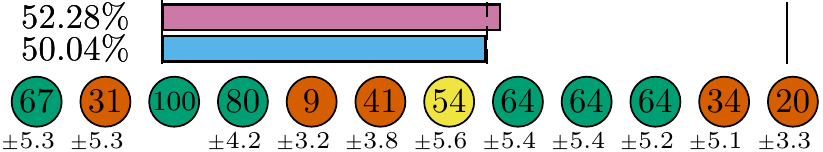} \\ \hline
\notbot{\semisplitV{S-nodal,\ravemix}{S,\mastmix}{Mod}}    & \colG{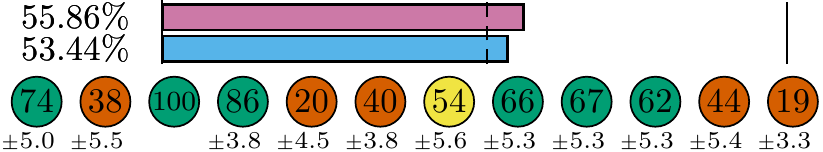} \\ \hline
\notbot{\semisplitV{S-nodal}{S,\mastsplit}{Mod}}    & \colG{M_Sn_SMs} \\ \hline
\notbot{\semisplitV{S-nodal}{S,\mastcontext}{Mod}}    & \colG{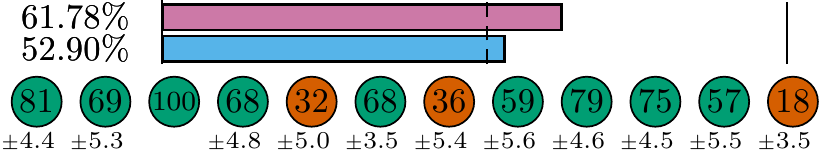} \\ \hline
\notbot{\semisplitV{S-nodal}{S,\mastmix}{Mod}}    & \colG{M_Sn_SMm7} \\ \hline
\notbot{\semisplitV{S-nodal}{S}{Mod}}    & \colG{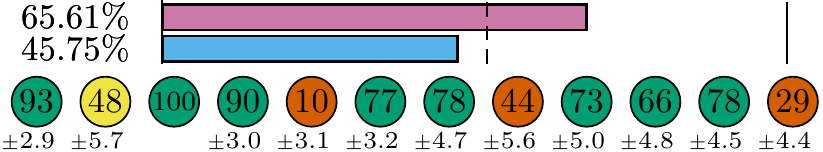} \\ \hline
&  \includegraphics[angle=90,width=\wid]{figures_winrates/gamenames.pdf}\\\hline
\end{tabular}\end{center}
\end{table}
\begin{table}[!htb]\small\renewcommand{\arraystretch}{0.1}
\newcommand{\colTwo}[1]{\multicolumn{2}{c|}{#1}}
\newcommand{\colBTwo}[1]{\multicolumn{2}{|c|}{#1}}
\newcommand{\wid}[0]{210px}
\newcommand{\colG}[1]{\includegraphics[width=\wid]{figures_winrates/#1.pdf}}
\newcommand{\notbot}[1]{\vbox{\hbox{\strut #1}\hbox{\strut \vphantom{a}} }}
\caption{Win rates of semisplit agents based on \semisplit{O}{S}{Mod} with variants of action-based heuristics.}\label{tab:agents_heur2}
\begin{center}\begin{tabular}{|l|c|}\hline
\multicolumn{1}{|c|}{Agent}     &  Winrates vs. \orthodoxMR\phantom{\LARGE X} \\[4pt]\hline
\notbot{\semisplitV{O}{S,\mastsplit}{Mod}}    & \colG{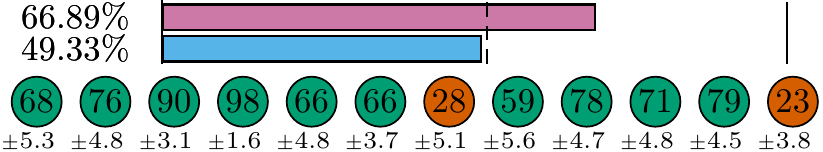} \\ \hline
\notbot{\semisplitV{O,\ravejoin}{S,\mastsplit}{Mod}}  & \colG{M_ORj_SMs} \\ \hline
\notbot{\semisplitV{O,\ravejoin}{S,\mastcontext}{Mod}}    & \colG{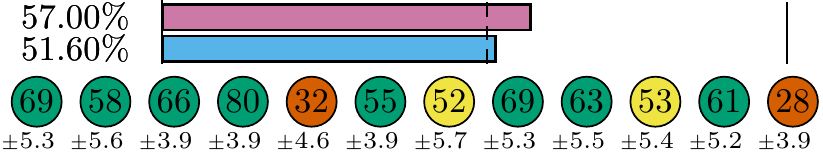} \\ \hline
\notbot{\semisplitV{O,\ravejoin}{S,\mastmix}{Mod}}   & \colG{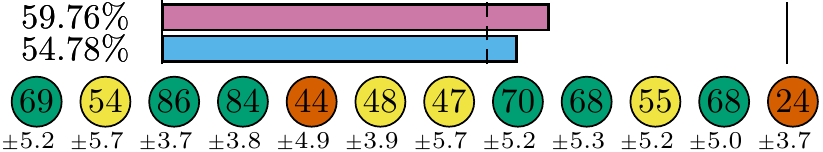} \\ \hline
&  \includegraphics[angle=90,width=\wid]{figures_winrates/gamenames.pdf}\\\hline
\end{tabular}\end{center}
\end{table}
\begin{table}[!htb]\small\renewcommand{\arraystretch}{0.1}
\newcommand{\colTwo}[1]{\multicolumn{2}{c|}{#1}}
\newcommand{\colBTwo}[1]{\multicolumn{2}{|c|}{#1}}
\newcommand{\wid}[0]{210px}
\newcommand{\colG}[1]{\includegraphics[width=\wid]{figures_winrates/#1.pdf}}
\newcommand{\notbot}[1]{\vbox{\hbox{\strut #1}\hbox{\strut \vphantom{a}} }} \caption{Win rates of semisplit agents based on \semisplit{R-nodal}{S}{Mod} with variants of action-based heuristics.}\label{tab:agents_heur3}
\begin{center}\begin{tabular}{|l|c|}\hline
\multicolumn{1}{|c|}{Agent}     &  Winrates vs. \orthodoxMR\phantom{\LARGE X} \\[4pt]\hline
\notbot{\semisplitV{R-nodal}{S,\mastsplit}{Mod}}    & \colG{M_Rn_SMs} \\ \hline
\notbot{\semisplitV{R-nodal}{S,\mastmix}{Mod}}   & \colG{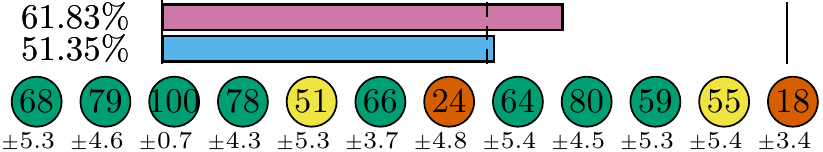} \\ \hline
\notbot{\semisplitV{R-nodal,\ravecontext}{S,\mastsplit}{Mod}}    & \colG{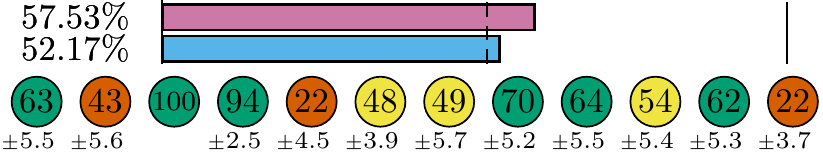} \\ \hline
&  \includegraphics[angle=90,width=\wid]{figures_winrates/gamenames.pdf}\\\hline
\end{tabular}\end{center}
\end{table}


\end{document}